\documentclass{article}

\usepackage[preprint]{neurips_2023}

\usepackage[utf8]{inputenc} 
\usepackage[T1]{fontenc}    
\usepackage{CJKutf8}
\usepackage{hyperref}       
\usepackage{url}            
\usepackage{booktabs}       
\usepackage{amsfonts}       

\usepackage{nicefrac}       
\usepackage{microtype}      
\usepackage[table,xcdraw]{xcolor}         

\usepackage{subcaption}
\usepackage{wrapfig}
\usepackage{multirow}
\usepackage{makecell}
\usepackage{xspace}
\usepackage{enumitem}
\usepackage{amsmath}
\usepackage{listings}
\usepackage{algorithm}
\usepackage{algorithmic}
\usepackage{amsfonts,amssymb}
\usepackage{amsthm}
\usepackage{mathrsfs}
\usepackage{subcaption}
\usepackage{natbib}
\setcitestyle{numbers,square}
\usepackage{graphicx}       
\usepackage{cleveref}
\usepackage{fontawesome}
\usepackage{utfsym}
\usepackage[most]{tcolorbox}
\usepackage{siunitx} 
\usepackage{tabularx}
\usepackage{CJKutf8}
\usepackage{fancyvrb}
\usepackage{colortbl}
\usepackage{array}
\usepackage{threeparttable}
\usepackage{CJKutf8}
\usepackage{CJK}
\usepackage{booktabs}
\usepackage[table]{xcolor}
\usepackage{lipsum}
\usepackage{listings}
\definecolor{demphcolor}{RGB}{144,144,144}

\newcommand{\hhide}[1]{}
\newcommand{\hide}[1]{}

\definecolor{zhipublue}{HTML}{3859FF}

\newtcolorbox{promptbox}[1][]{
  breakable,
  title=#1,
  colback=gray!5,
  colframe=black,
  colbacktitle=gray!15,
  coltitle=black,
  fonttitle=\bfseries,
  bottomrule=1.5pt,
  toprule=1.5pt,
  leftrule=1pt,
  rightrule=1pt,
  arc=0pt,
  outer arc=0pt,
  enhanced,
  before upper={\parindent=1.5em} 
}

\theoremstyle{definition}

\title{GLM-4.5: Agentic, Reasoning, and Coding (ARC) Foundation Models}

 \author{
{GLM-4.5 Team}
~\\\\
Zhipu AI ~\&~ Tsinghua University\\\\
(For the complete list of authors, please refer to the \hyperref[sec:contribution]{Contribution} section)
{}
}

\begin{document}

\maketitle

\vspace{-1.5em}
\begin{abstract}

\vspace{-0.5em}
We present GLM-4.5, an open-source Mixture-of-Experts (MoE) large language model with 355B total parameters and 32B activated parameters, featuring a hybrid reasoning method that supports both thinking and direct response modes. Through multi-stage training on 23T tokens and comprehensive post-training with expert model iteration and reinforcement learning, GLM-4.5 achieves strong performance across agentic, reasoning, and coding (ARC) tasks, scoring 70.1\% on TAU-Bench, 91.0\% on AIME 24, and 64.2\% on SWE-bench Verified. With much fewer parameters than several competitors, GLM-4.5 ranks 3rd overall among all evaluated models and 2nd on agentic benchmarks. We release both GLM-4.5 (355B parameters) and a compact version, GLM-4.5-Air (106B parameters), to advance research in reasoning and agentic AI systems. Code, models, and more information are available at \url{https://github.com/zai-org/GLM-4.5}.


     

\end{abstract}

\vspace{-1.5em}
\begin{figure}[H]
    \centering
    \includegraphics[width=0.925\linewidth]{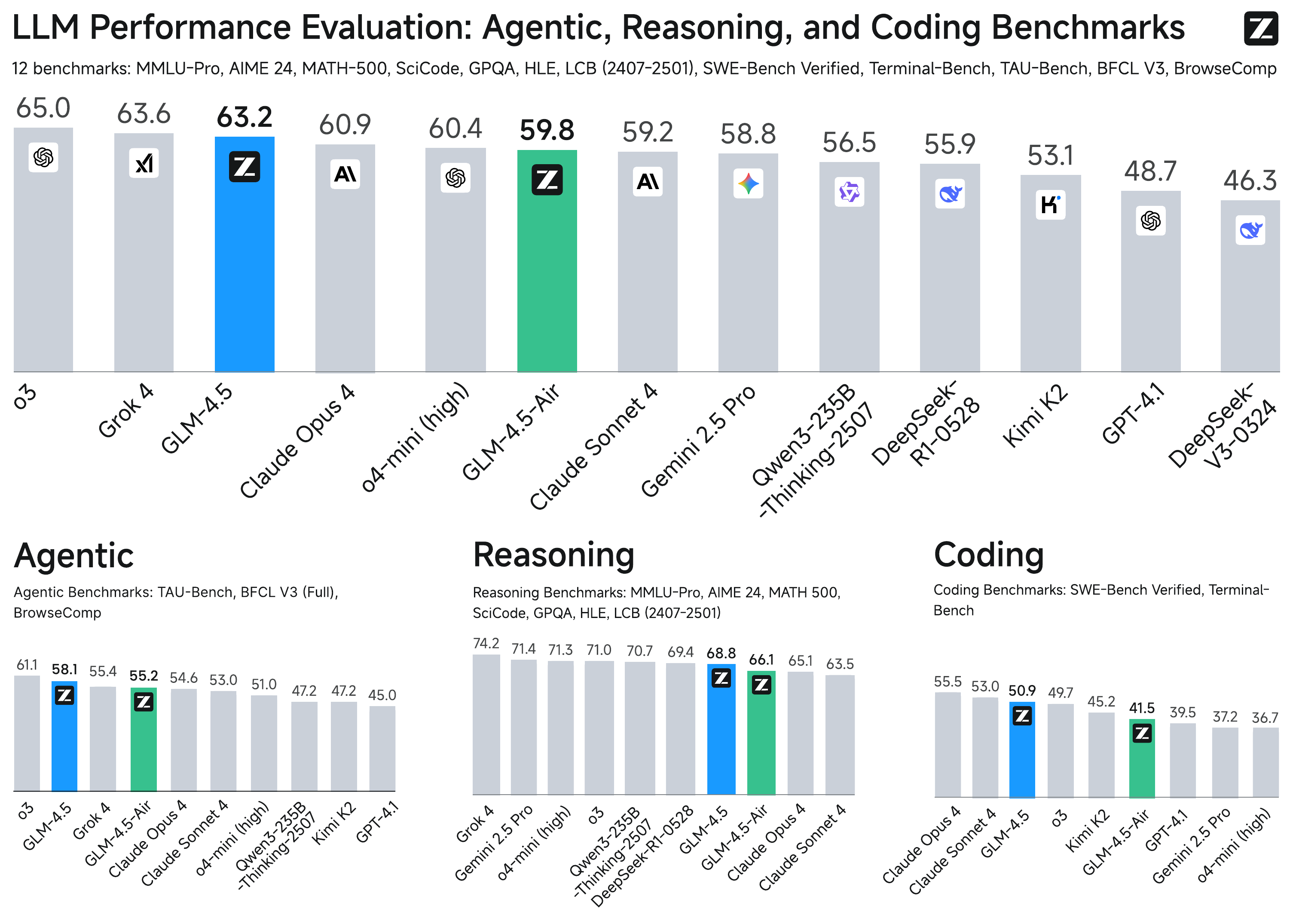}
    \caption{
    Average performance on agentic, reasoning, and coding (ARC) benchmarks. Overall, GLM-4.5 achieves a rank of 3rd, with GLM-4.5-Air following at rank 6th.
    The models listed are evaluated as of July 28, 2025.
    }
    \label{fig:arc_benchmark}
\end{figure}

\section{Introduction}

Large language models (LLMs) are rapidly evolving from general knowledge repositories~\cite{brown2020language,touvron2023llama,zengglm,team2023gemini,liu2024deepseek} into general problem-solvers. The ultimate ambition, often associated with Artificial General Intelligence (AGI), is to create models with human-level cognitive capabilities across diverse domains. This requires a unified mastery of complex problem-solving, generalization, and self-improvement, moving beyond task-specific excellence.

As LLMs become more integrated into real-world scenarios, the key to enhancing actual productivity and solving complex professional tasks lies in developing specific core capabilities.
We identify three critical, interconnected capabilities as the measure of a truly generalist model: \textbf{Agentic} abilities for interacting with external tools and the real world; complex \textbf{Reasoning} for solving multi-step problems in domains like mathematics and science; and advanced \textbf{Coding} skills for tackling real-world software engineering tasks. While state-of-the-art proprietary models like OpenAI's o1/o3~\cite{jaech2024openai} and Anthropic's Claude Sonnet 4 have demonstrated groundbreaking performance in specific ARC domains (e.g., mathematical reasoning or code fixing~\cite{jimenez2023swe}), a single, powerful open-source model that excels across all three areas has remained elusive.

This paper introduces two new models: GLM-4.5 and GLM-4.5-Air, toward the goal of unifying all the different capabilities.
The new models outperform existing open-source LLM models~\cite{guo2025deepseek,team2025kimi,yang2025qwen3} across the board, with significant gains in agentic, reasoning, and coding tasks. GLM-4.5 and GLM-4.5-Air both feature hybrid reasoning modes: thinking mode for complex reasoning and agentic tasks, and non-thinking mode for instant responses. GLM-4.5 is our first MoE model, with 355B total parameters and 32B activated parameters. GLM-4.5 demonstrates strong performance on the following ARC benchmarks:

\begin{itemize}[leftmargin=*,itemsep=0pt,parsep=0.4em,topsep=0.0em,partopsep=0.0em]
     \item \textbf{Agentic:} GLM-4.5 scores 70.1\% on TAU-Bench and 77.8\% on BFCL v3~\cite{patil2025bfcl}, on par with Claude Sonnet 4. For web browsing agents, GLM-4.5 scores 26.4\% on BrowseComp~\cite{wei2025browsecomp}, clearly outperforming Claude Opus 4 (18.8\%) and close to o4-mini-high (28.3\%).
     
    \item \textbf{Reasoning:} GLM-4.5 demonstrates outstanding performance on a suite of challenging reasoning benchmarks, achieving 91.0\% on AIME 24, 79.1\% on GPQA~\cite{rein2024gpqa}, 72.9\% on LiveCodeBench (2407-2501)~\cite{jainlivecodebench}, and 14.4\% on HLE (Humanity's Last Exam)~\cite{phan2025humanity}.
    
    \item \textbf{Coding:} GLM-4.5 scores 64.2\% on SWE-bench Verified~\cite{jimenez2023swe} and 37.5\% on Terminal-Bench~\cite{tbench_2025}, outperforming GPT-4.1 and Gemini-2.5-pro, close to Claude Sonnet 4.
\end{itemize}

GLM-4.5-Air is a smaller MoE model with 106B parameters. It represents a significant leap among models at the 100B scale, matching or exceeding Qwen3-235B-A22B~\cite{yang2025qwen3} and MiniMax-M1~\cite{chen2025minimax}.

\begin{figure}
    \centering
    \includegraphics[width=0.9\linewidth, trim=0 15 0 15, clip]{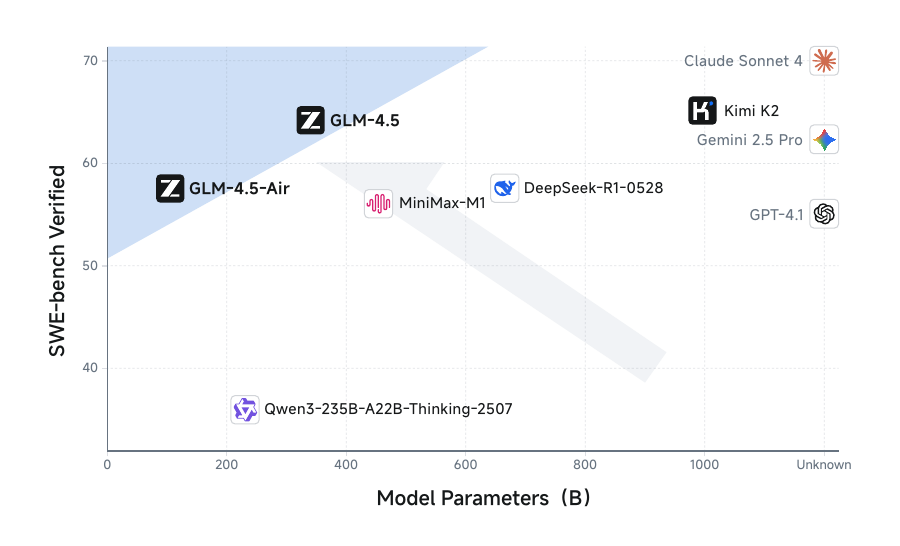}
    \caption{SWE-bench verified scores vs model parameters. Proprietary models are listed as unknown at the right side.}
    \label{fig:swe-parameter}
\end{figure}

In \Cref{fig:arc_benchmark}, we show the average performance on 12 benchmarks across agentic, reasoning, and coding (ARC) tasks. Overall, GLM-4.5 is ranked in the \textbf{3rd} place and GLM-4.5-Air is ranked in the \textbf{6th}. On agentic tasks, GLM-4.5 is ranked in the 2nd place, following OpenAI o3. On coding tasks, GLM-4.5 is ranked in the third place, close to Claude Sonnet 4. Note that GLM-4.5 is highly parameter-efficient, with only half the parameters of DeepSeek-R1~\cite{guo2025deepseek} and one-third those of Kimi K2~\cite{team2025kimi}. In \Cref{fig:swe-parameter}, we report the scores on SWE-bench Verified vs model parameters of different open-source models, where GLM-4.5 and GLM-4.5-Air lie on the Pareto Frontier.
More evaluation results are detailed in~\Cref{sec:evaluation}.

Both GLM-4.5 and GLM-4.5-Air are available on \url{Z.ai}, \url{BigModel.cn}, and also as open-source models on \url{https://huggingface.co/zai-org/GLM-4.5}. We also open-sourced an evaluation toolkit at \url{https://github.com/zai-org/glm-simple-evals} to ensure the reproducibility of our benchmark results.

\section{Pre-Training}
 
\subsection{Architecture}

In the GLM-4.5 series, we adopt the MoE architecture, which improves the computational efficiency of both training and inference. We employ loss-free balance routing~\cite{wang2024auxiliary} and sigmoid gates for MoE layers~\cite{liu2024deepseek}. Different from DeepSeek-V3~\cite{liu2024deepseek} and Kimi K2~\cite{team2025kimi}, we reduce the width (hidden dimension and number of routed experts) of the model and increase its height (number of layers), as we found that deeper models exhibited better reasoning capacity. In the self-attention component, we employ Grouped-Query Attention with partial RoPE. Furthermore, we utilize 2.5 times more attention heads (96 heads for a 5120 hidden dimension). Counterintuitively, while this increased head count does not improve training loss compared to models with fewer heads, it consistently improves performance on reasoning benchmarks such as MMLU and BBH. We also incorporate QK-Norm~\cite{henry2020querykeynormalizationtransformers} to stabilize the range of attention logits. For both GLM-4.5 and GLM-4.5-Air, we add an MoE layer as the MTP (Multi-Token Prediction) layer~\cite{gloeckle2024mtp} to support speculative decoding during inference.

\begin{table}[!h]
    \centering
    \caption{Model architecture of GLM-4.5 and GLM-4.5-Air. When counting parameters, for GLM-4.5 and GLM-4.5-Air, we include the parameters of MTP layers but not word embeddings and the output layer.}
    
    \label{tab:placeholder}
    \begin{tabular}{l|cc|cc}
    \toprule
       Model & \textbf{GLM-4.5} & \textbf{GLM-4.5-Air} & \textbf{DeepSeek-V3} & \textbf{Kimi K2} \\
    \midrule
       \# Total Parameters & 355B & 106B & 671B & 1043B \\
       \# Activated Parameters & 32B & 12B & 37B & 32B \\
       \# Dense Layers  & 3 & 1 & 3 & 1 \\
       \# MoE Layers & 89 & 45 & 58 & 60 \\
       \# MTP Layers & 1 & 1 & 1 & 0 \\
       Hidden Dim & 5120 & 4096 & 7168 & 7168 \\
       Dense Intermediate Dim & 12288 & 10944 & 18432 & 18432 \\
       MoE Intermediate Dim & 1536 & 1408 & 2048 & 2048 \\
       Attention Head Dim & 128 & 128 & 192 & 192\\
       \# Attention Heads & 96 & 96 & 128 & 64 \\
       \# Key-Value Heads & 8 & 8 & 128 & 64 \\
       \# Experts (total) & 160 & 128 & 256 & 384 \\
       \# Experts Active Per Token & 8 & 8 & 8 & 8\\
       \# Shared Experts & 1 & 1 & 1 & 1\\
       QK-Norm & Yes & No & No & No \\
    \bottomrule
    \end{tabular}
\end{table}

\subsection{Pre-Training Data}
Our pre-training corpus includes documents from webpages, social media, books, papers, and code repositories. We carefully design the data processing pipelines for different sources. 

\paragraph{Web} The majority of our pre-training documents are English and Chinese webpages crawled from the Internet. Inspired by Nemotron-CC~\cite{su2024nemotron}, we divide the crawled webpages into buckets of different quality scores. We up-sample documents from the bucket with higher quality scores
and discard documents from the bucket with the lowest quality scores. The bucket with the highest quality scores contributes over 3.2 epochs during pre-training. In this way, the pre-training corpus can emphasize the high-frequency knowledge for reasoning tasks and also improve coverage for long-tail world knowledge. We have also found a large number of similar webpages automatically generated from templates and assigned high scores. Such webpages cannot be removed by MinHash deduplication. We additionally apply the SemDedup~\cite{abbas2023semdedup} pipeline to remove those similar webpages based on document embeddings.

\paragraph{Multilingual} To support more natural languages, we include multilingual documents in our pre-training corpus. The multilingual corpus comes from both our crawled webpages and Fineweb-2~\cite{penedo2025fineweb2}. We apply a quality classifier that judges the educational utility of documents and up-sample high-quality multilingual documents.

\paragraph{Code} We curated source code data from GitHub and various code hosting platforms. The code corpus undergoes a preliminary rule-based filtering, followed by classification using language-specific quality models that categorize samples into three tiers: high-quality, medium-quality, and low-quality. During training, we up-sampled high-quality code while excluding low-quality samples. Moreover, the Fill-In-the-Middle~\cite{bavarian2022efficienttraininglanguagemodels} training objective is applied to all source code data. 
For code-related web documents, we employ a two-stage retrieval process from our text pre-training corpus. Documents are initially selected based on two criteria: presence of HTML code tags, or identification by a FastText~\citep{joulin2017bag} classifier trained to detect code-related content. Subsequently, the retrieved documents undergo quality assessment using a dedicated model that classifies them into high-, medium-, or low-quality categories, following the same quality-based sampling strategy for source code. Finally, a fine-grained parser is employed to re-parse the selected web pages to better preserve the formats and contents of the code.

\paragraph{Math \& Science} To enhance the reasoning capacity, we collect documents related to mathematics and science from webpages, books, and papers. We apply a large language model to score candidate documents based on the ratio of educational content about mathematics and science, and train a small-scale classifier to predict the scores. Documents in the pre-training corpus with scores above a certain threshold are up-sampled.

The pre-training process of GLM-4.5 is divided into two stages. In the first stage, the model is mainly trained on general documents from webpages. During the second stage, we up-sample the source code from GitHub and webpages related to coding, mathematics, and science.

\subsection{Mid-Training: Boost Reasoning \& Agentic Capacity}

\begin{figure}
    \centering
    \includegraphics[width=\linewidth]{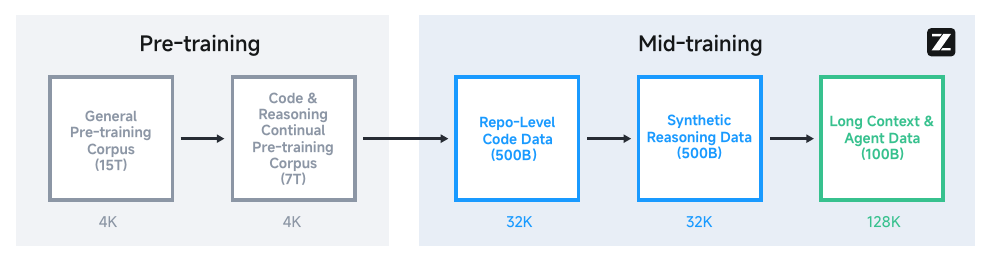}
    \caption{Pre-training and mid-training stages for GLM-4.5. We adapt a multi-stage training recipe and extend the sequence length from 4K to 128K.}
    \label{fig:pretrain}
\end{figure}

After pre-training, we add several stages to further boost the model's performance on important application areas. Unlike traditional pre-training on large-scale general documents, these training stages utilize medium-size domain-specific datasets, including instruction data. Therefore, we denote these training stages as mid-training, which includes the following.

\paragraph{Repo-level Code Training} At this training stage, we add concatenated code files from the same repository to learn cross-file dependency. To improve the model's software engineering capability, we also include model-filtered issues, pull requests (PRs), and commits from GitHub, with related issues, PRs, and commits concatenated into one context and commits organized in a diff-like format. We extend the training sequence length from 4K to 32K to incorporate large repositories.

\paragraph{Synthetic Reasoning Data Training} At this stage, we add synthetic reasoning content for math, science, and coding competitions. We collect a large number of questions and answers related to the reasoning tasks from webpages and books, and synthesize reasoning processes with a reasoning model.
\paragraph{Long-context \& Agent Training} To further boost the model's long-context performance, we extend the training sequence length from 32K to 128K and up-sample long documents from the pre-training corpus. Large-scale synthetic agent trajectories are also incorporated at this stage.

In \Cref{fig:pretrain}, we show the complete stages for pre-training and mid-training. The maximum sequence length is kept at 4,096 in pre-training, and is extended from 32,768 to 131,072 in mid-training. During pre-training, we did not use best-fit packing~\cite{ding2024fewer} since random truncation is a good data-augmentation strategy for pre-training documents. For datasets in mid-training, we applied best-fit packing to avoid truncating the reasoning process or repo-level code.

\subsection{Hyper-Parameters}

We employed the Muon optimizer~\cite{jordan6muon,liu2025muon} for all parameters except word embedding, bias, and weights for RMSNorm. For hyperparameters, we set the Newton-Schulz iteration steps $N$ to 5, momentum $\mu$ to 0.95, and scaled Muon's update RMS to 0.2. We observed that the Muon optimizer can accelerate convergence and tolerate larger batch sizes. We used cosine decay schedule for learning rate, instead of warmup-stable-decay (WSD) schedule~\cite{huminicpm}. Our early experiments showed that models trained with the WSD schedule perform worse on general benchmarks (SimpleQA, MMLU), indicating underfitting in the stable stage. The learning rate went through a warm-up stage from 0 to 2.5e-4 and a decaying stage to 2.5e-5 until the end of mid-training. We used a batch size warmup strategy, where the batch size was gradually increased from 16M tokens to 64M tokens in the training of the first 500B tokens, and remained constant in the remaining of training. 
For regularization, we set the weight decay ratio to 0.1 and did not use dropout. We set the maximum sequence length to 4,096 during pre-training, and extended it to 32,768 and 131,072 during the mid-training stage as shown in~\Cref{fig:pretrain}. When extending the sequence length to 32K, we also adjusted RoPE's base frequency from 10,000 to 1,000,000 for better long-context modeling ability. For loss-free balance routing, we set the bias update rate to 0.001 for the first 15T tokens, and to 0.0 for the remaining tokens. We also applied auxiliary sequence-level balance loss with a 0.0001 weight to avoid extreme imbalance within any single sequence. The MTP loss weight $\lambda$ was set to 0.3 for the first 15T tokens, and to 0.1 for the remaining tokens.

\section{Post-Training: Expert Model Iteration}

We divide the post-training process into two distinct stages. In stage 1 (\emph{Expert Training}), we construct expert models specializing in three domains: Reasoning, Agent, and General chat. In stage 2 (\emph{Unified Training}), we employ self-distillation techniques to integrate multiple experts, ultimately delivering a comprehensive model capable of generating responses through both deliberative reasoning and direct response modes.

\subsection{Supervised Fine-Tuning}

We perform Supervised Fine-Tuning (SFT) at the beginning of both Stage 1 (\emph{Expert Training}) and Stage 2 (\emph{Unified Training}). In the expert training stage, the primary role of SFT is to provide a cold start, empowering the model with basic chat, reasoning, and tool-use capabilities, which can then be further enhanced in subsequent expert RL training to achieve improved performance. In the unified training stage, the purpose of SFT is to distill the capabilities of different expert models into one hybrid reasoning generalist capable of handling different types of tasks.

\paragraph{Cold Start SFT} During the cold-start phase, we utilize a small set of supervised fine-tuning (SFT) data with extended Chain-of-Thought (CoT) responses. This approach ensures that each expert model possesses adequate foundational ability prior to the reinforcement learning phase.

\paragraph{Overall SFT} In the Overall SFT stage, we collect millions of samples covering reasoning tasks (math, code, science, etc.), general chat (writing, translation, summarization, chit chat, etc.), agentic tasks (basic tool using, coding ability especially for authentic project development, etc.), and long-context understanding tasks from the previously trained expert models, and train the base model with a maximum context length of 128K tokens. By distilling from the output of distinct experts, the model learns to apply the most effective long CoT reasoning for each task to arrive at accurate answers. Especially, recognizing that a prolonged thinking process is unnecessary for certain domains that demand quick responses (such as chit chat), we meticulously balanced training data containing full reasoning with data lacking explicit thought processes. This approach allows the model to operate in both the reflective and immediate response modes, thereby creating a \emph{hybrid reasoning model}. Moreover, we find the following strategy helpful in preparing SFT data to derive optimal performance.

\paragraph{Reducing Character Escaping in Function Call Templates} Although function call parameters are predominantly represented in JSON format in contemporary implementations, a significant challenge emerges when these parameters contain code segments. In such cases, a substantial proportion of characters within the code require escaping, compelling the model to generate extensive escape characters, thereby increasing the learning burden for the model. While this issue poses minimal concern for models primarily designed for general chat, it represents a non-trivial challenge for agentic foundation models where function calling is a core capability.
To mitigate this limitation, we propose a novel function call template that encapsulates function call keys and values within XML-like special token tags. This approach substantially reduces the necessity for character escaping in code segments, as the vast majority of code can be represented in its native form without escaping. Experimental results demonstrate that the proposed function call template does not compromise the performance of function call execution while recucing escaping. The following example (Figure \ref{fig:fc-template}) illustrates the structure of our proposed function call template. Detailed code implementation can be found in our open-source repository.

\begin{tcolorbox}[left=0mm,right=0mm,top=0mm,bottom=0mm,boxsep=1mm,arc=0mm,boxrule=0pt, frame empty, breakable]
\scriptsize
\begin{lstlisting}
<|system|>
# Tools

You may call one or more functions to assist with the user query.

You are provided with function signatures within <tools></tools> XML tags:
<tools>
{"name": "get_weather", "description": "Get the weather of a city for a specific date.", "parameters": {"type": "object", "properties": {"city": {"type": "string", "description": "The city to get weather for, in Chinese."}, "date": {"type": "string", "description": "The date in YYYY-MM-DD format."}}, "required": ["city"]}}
</tools>

For each function call, output the function name and arguments within the following XML format:
<tool_call>{function-name}
<arg_key>{arg-key-1}</arg_key>
<arg_value>{arg-value-1}</arg_value>
<arg_key>{arg-key-2}</arg_key>
<arg_value>{arg-value-2}</arg_value>
...
</tool_call><|system|>
You are a helpful assistant.<|user|>
Today is June 26, 2024. Could you please check the weather in Beijing and Shanghai for tomorrow<|assistant|>
<think>The user wants to check the weather of Beijing and Shanghai tomorrow. I need to call the get_weather function respectively to check Beijing and Shanghai.</think>
I will call the get_weather function to check the weather in Beijing and Shanghai.
<tool_call>get_weather
<arg_key>city</arg_key>
<arg_value>Beijing</arg_value>
<arg_key>date</arg_key>
<arg_value>2024-06-27</arg_value>
</tool_call>
<tool_call>get_weather
<arg_key>city</arg_key>
<arg_value>Shanghai</arg_value>
<arg_key>date</arg_key>
<arg_value>2024-06-27</arg_value>
</tool_call><|observation|>
<tool_response>
{"city": "Beijing", "date": "2024-06-27", "weather": "Sunny", "temperature": "26C"}
</tool_response>
<tool_response>
{"city": "Shanghai", "date": "2024-06-27", "weather": "Overcast", "temperature": "29C"}
</tool_response><|assistant|>
<think>I have obtained the weather query results of get_weather for Beijing and Shanghai respectively and can reply to users directly.</think>
It will be sunny in Beijing tomorrow with a temperature of 26 degrees Celsius. The weather in Shanghai is overcast with a temperature of 29 degrees Celsius.<|user|>
\end{lstlisting}
\end{tcolorbox}
\noindent\begin{minipage}{\textwidth}
\captionof{figure}{One example of function call template.}
\label{fig:fc-template}
\end{minipage}

\paragraph{Rejection Sampling} When sampling from expert models, we employ a comprehensive multistage filtering pipeline that includes: (1) removing repetitive, excessively short, or truncated samples, as well as those that fail to conform to valid reasoning formats; (2) conducting correctness verification for samples with objective answers; (3) utilizing reward models to filter responses to subjective questions; and (4) for tool-calling scenarios, ensuring adherence to proper tool invocation protocols and verification that trajectories reach the expected terminal states.

\paragraph{Prompt Selection and Response-Level Scaling} Filtering challenging prompts and conducting response scaling on them prove to be effective. We experimented with removing the prompts in the bottom 50\% based on response lengths, resulting in a 2\%-4\% improvement in math and science tasks, despite training with only half the data. Notably, we found that applying response scaling to these hard prompts can lead to further gains. Generating four responses for each prompt brought an additional 1\%-2\% improvement.

\paragraph{Automatic Agentic SFT Data Construction} The construction of agentic SFT data involves four steps: 1. \emph{Agentic Framework and Tool Collection}: We gather a set of agentic frameworks and real-world tool APIs and MCP servers, while also leveraging LLMs to automatically construct and simulate a batch of tools. 2. \emph{Task Synthesis}: Based on these frameworks and tools, we automatically synthesize a collection of agentic tasks. On the one hand, for relatively mature frameworks, we leverage LLMs to comprehend their functionalities and automatically generate relevant queries or tasks. On the other hand, for more fragmented or disparate tools, we first select a representative subset and similarly employ LLMs to construct tasks about this subset. These tasks encompass both single-step and multi-step tool calling scenarios. 3. \emph{Trajectory Generation}: For each synthesized task, we utilize existing LLMs to generate tool-call trajectories. Additionally, by employing the LLM as a user simulator, multi-step tool-call tasks are converted into trajectories involving multiple rounds of dialogue. 4. \emph{Quality Filtering}: For each trajectory, multiple judge agents are used to evaluate whether the task is completed. Only successful trajectories are retained.




\subsection{Reasoning RL}

Reasoning RL focuses on enhancing a model's capabilities in domains that demand logical deduction, structured problem-solving, and verifiable accuracy. This includes critical areas such as mathematics, code generation, and scientific reasoning. A defining characteristic of these tasks is the high precision of their reward signals, as correctness can often be determined programmatically or with objective clarity. Mastery in these areas is not only crucial for advancing the raw intelligence of models but also serves as a fundamental building block for more complex, multi-step agentic behaviors. Recognizing the unique challenges and opportunities within reasoning RL, we have developed a suite of specialized techniques to effectively train our models. These methods, detailed below, are designed to address issues such as training efficiency, sample diversity, and data quality. Our overall RL algorithm builds upon the GRPO~\cite{shao2024deepseekmath} framework, excluding the KL loss term. The comparison curves shown in this section are based on our smaller experimental model, not on GLM-4.5.

\begin{figure}[t]
    \centering
    \includegraphics[width=0.85\textwidth]{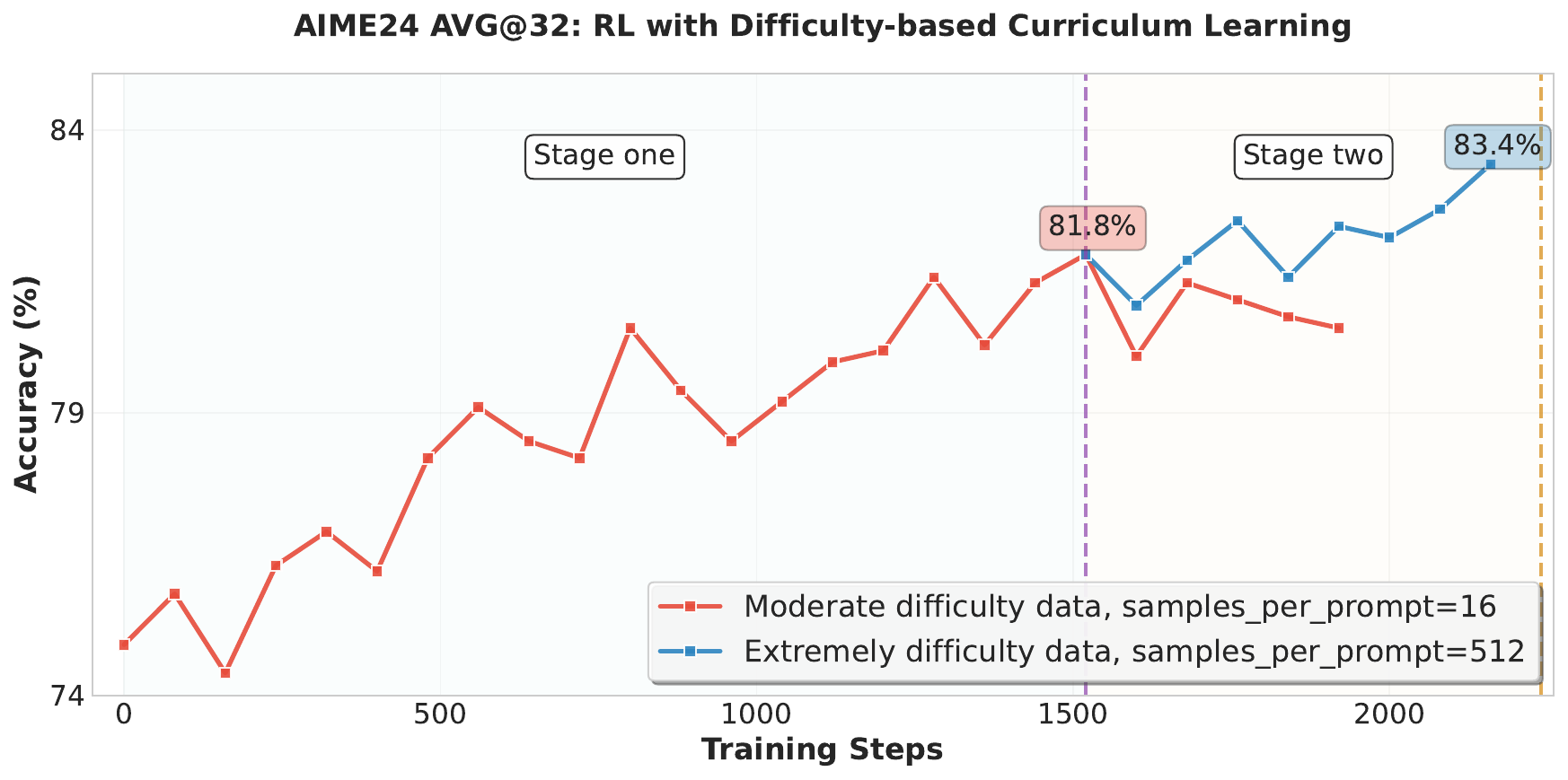}
    \caption{\textbf{Effectiveness of the two-stage difficulty-based curriculum on AIME'24.} The \textcolor{blue}{blue line} (our method) switches to extremely difficult problems (pass@8$=$0, pass@512$>$0) in the second stage, showing continued improvement. The \textcolor{red}{red line} (baseline) continues with moderate-difficulty problems and plateaus.}
    \label{fig:curriculum_rl}
\end{figure}

\paragraph{Difficulty-based Curriculum Learning}
During reinforcement learning, the model's proficiency evolves, creating a mismatch with static training data. In the later stages, as the model becomes more capable, overly simple data can lead to rollouts where all rewards are 1s. Conversely, in the early stages, excessively difficult data often results in batches where all rewards are 0s. In both scenarios, the lack of reward variance provides no useful gradient signal, severely hindering training efficiency. To address this challenge, we employ a two-stage difficulty-based curriculum for RL. The effectiveness of this and other strategies discussed below is validated through controlled experiments on a smaller model, which allows for rapid iteration and precise ablation studies. As shown in Figure~\ref{fig:curriculum_rl}, this two-stage approach enables the model to consistently surpass its performance ceiling. Crucially, to maintain high signal quality and reduce noise, all problems used in the second stage are strictly sourced from a pool with verified correct answers.

\begin{figure}[t]
    \centering
    \includegraphics[width=1.0\textwidth]{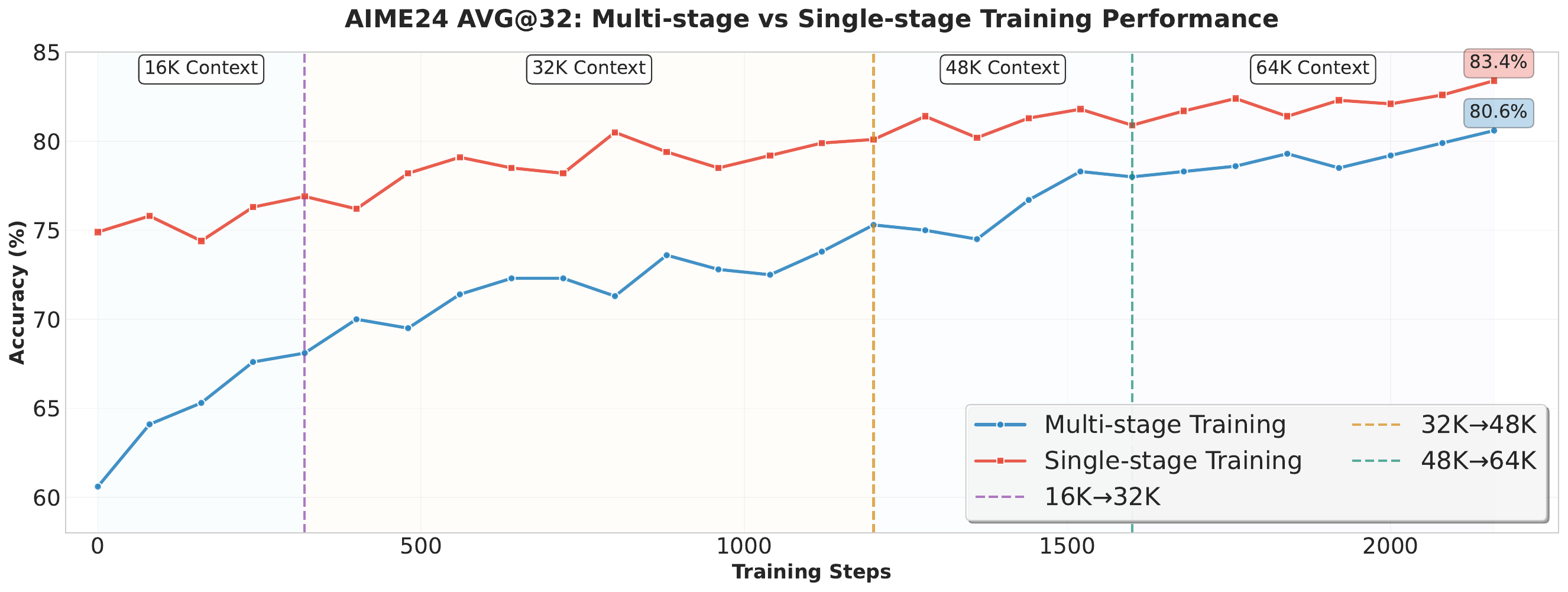}
    \caption{\textbf{Single-stage vs. multi-stage RL at 64K context length.} The \textcolor{red}{red line} (single-stage at 64K) achieves superior performance. The \textcolor{blue}{blue line} (multi-stage with progressively increasing length) suffers from an irreversible performance drop in early stages, limiting its final performance.}
    \label{fig:single_stage_rl}
\end{figure}

\paragraph{Single-Stage RL at 64K Output Length}
Previous research~\cite{deepscaler2025} has suggested conducting RL in multiple stages with progressively increasing maximum output lengths. However, our experiments reveal that this multi-stage approach is less effective than a single-stage RL process conducted directly at the maximum target length of 64K. Since the initial Supervised Fine-Tuning (SFT) has already conditioned the model on generating 64K-length responses, introducing RL stages with shorter maximum lengths can cause the model to ``unlearn'' its long-context capabilities. This often leads to a significant and irreversible drop in performance, as the model's average output length decreases. This degradation is difficult to recover from in the final 64K-length RL stage, thus limiting further improvement. Our experiments confirm this observation: as demonstrated in Figure~\ref{fig:single_stage_rl}, applying RL directly at the full 64K-length continually pushes the model's limits and yields better performance.

\paragraph{Dynamic Sampling Temperature}
During RL, the sampling temperature is a key parameter for controlling trajectory diversity. A temperature that is too low leads to convergent, less exploratory outputs, while one that is too high introduces low-quality, noisy samples, undermining both model accuracy and training efficiency. Using a fixed sampling temperature is suboptimal because it fails to adapt as the policy distribution becomes more concentrated (i.e., has lower entropy), often resulting in insufficient exploration at later stages. Therefore, we propose dynamically adjusting the sampling temperature to maintain a healthy balance between accuracy and exploration. Specifically, when the average reward of rollouts stabilizes, we identify this as a convergence phase and increase the sampling temperature to encourage greater diversity. To mitigate the risk of introducing excessive noise, we implement a quality-control mechanism: we periodically evaluate model performance on a held-out validation set across a range of temperatures. The temperature for the next training phase is then set to the maximum value that does not cause a performance drop of more than 1\% from the current optimum~\cite{Polaris2025}.

\begin{figure}[t]
    \centering
    \includegraphics[width=1.0\textwidth]{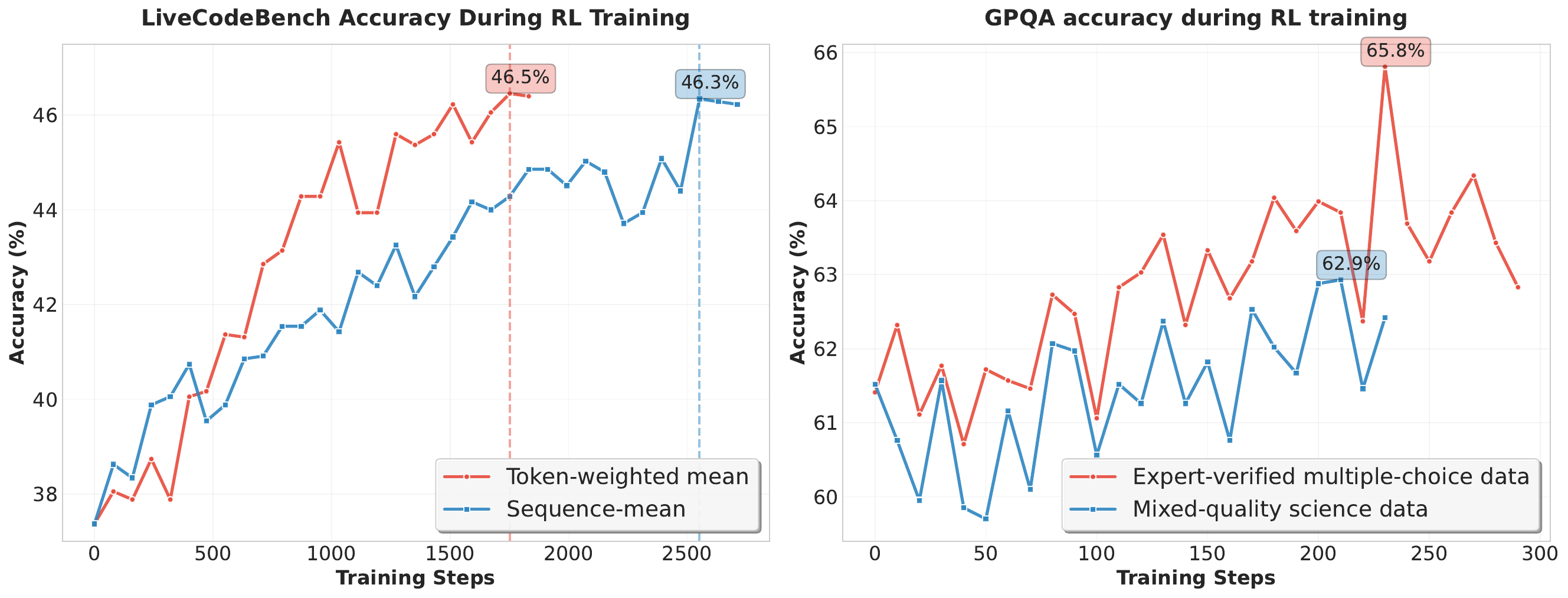}
    \caption{\textbf{Ablation Studies for Code and Science RL.} 
    \textbf{(Left)} Comparison of loss calculation methods for code RL. The \textcolor{red}{token-weighted mean loss} approach achieves faster convergence compared to the \textcolor{blue}{sequence-mean loss} baseline, accelerating the training process. 
    \textbf{(Right)} Ablation on data sources for science RL on the GPQA-Diamond benchmark. Training exclusively on a small set of high-quality, expert-verified multiple-choice questions yields the best performance, significantly outperforming training on mixed-quality data.}
    \label{fig:code_science_rl}
\end{figure}

\paragraph{Code and Science RL}
Compared to mathematics, RL for coding and scientific domains has received less attention in the literature. We conducted extensive controlled RL experiments in these areas and arrived at the following empirical conclusions. For \textbf{code RL}, we find that the choice of loss calculation is critical for training efficiency. As illustrated in Figure~\ref{fig:code_science_rl} (left), adopting a token-weighted mean loss is highly beneficial compared to a conventional sequence-mean loss. The token-weighted approach provides a finer-grained and more stable gradient signal, which leads to significantly faster convergence. This method also helps to alleviate the length bias inherent in sequence-level rewards and effectively suppresses the generation of overly simplistic or repetitive ``base case'' samples during training. For \textbf{science RL}, our findings on the GPQA-Diamond benchmark highlight that data quality and type are paramount factors. As shown in Figure~\ref{fig:code_science_rl} (right), using exclusively expert-verified multiple-choice questions for RL leads to significantly better performance compared to training with mixed-quality or unverified data. This result underscores that even for tasks with simple formats like multiple-choice, rigorously filtering the RL data pool to include only high-quality, challenging instances is crucial for effective model improvement.

\subsection{Agentic RL}

Reinforcement Learning from Human Feedback (RLHF) helps language models follow human instructions more faithfully. Applying RL to math and programming contests has further uncovered strong reasoning abilities and favorable scaling behavior on tasks whose outcomes can be objectively verified. Building on these insights, we focus on agentic settings—specifically web-search and code-generation agents—where every action or answer can be automatically checked. This built-in verifiability supplies dense, reliable rewards, enabling us to scale RL training more effectively.

\subsubsection{Data Collection and Synthesis for Agents}

For web‑search tasks and open‑domain information seeking, we develop a data‑synthesis pipeline that yields demanding question–answer pairs requiring multi‑step reasoning across multiple web sources. This corpus is designed to sharpen GLM’s ability to uncover elusive, interwoven facts on the internet. Dataset construction blends two approaches: (1) an automated pipeline powered by multi‑hop reasoning over knowledge graphs, and (2) human‑in‑the‑loop extraction and selective obfuscation of content from several web pages to prepare reinforcement‑learning training signals.

For software‑engineering tasks, we curate an extensive collection of GitHub pull requests and issues to create a realistic software‑development benchmark comprising user prompts and executable unit tests. All evaluations run inside a hardened sandbox with a distributed system, which provides both horizontal scalability and strong isolation guarantees.

\subsubsection{Pushing the Limits with Reinforcement Learning and Iterative Self-distillation}

We adopt the group-wise policy optimization algorithm for RL training.
For each problem \(x\), we sample \(K\) agent traces \(\{y_1,\dots,y_k\}\) from the previous policy
\(\pi_{\text{old}}\), and optimize the model \(\pi_\theta\) with respect to the following objective:
\[
L_{\text{RL}}(\theta)
   = \mathbb{E}_{x\sim\mathcal{D}}\!\left[
        \frac{1}{K}\sum_{i=1}^{K}
        \left(
            r(x,y_i) - \bar{r}(x)
        \right)
     \right],
\]
where $\bar{r}(x) \;=\; \frac{1}{k}\sum_{i=1}^{k} r\bigl(x,y_i\bigr)$
is the mean reward of the sampled responses. It is noted that only model-generated tokens are used for optimization, and the environment feedback is ignored in loss computation. 

\paragraph{Outcome Supervision with Process Action Format Penalty} For web search tasks, we use the accuracy of the final answer as a reward for the entire agent trace. For coding agents, we primarily utilize SWE data with verifiable test cases for RL training. Our experiments have shown that RL training on web search and SWE tasks leads to generalized performance improvements across other tasks and benchmarks, such as general tool usage and coding tasks like Terminal-Bench. Additionally, we apply a process format penalty to ensure the model generates correct tool call formats. If the model fails to produce the correct tool format during agent trace generation, the process will be halted, and the trace will receive a zero reward.

\paragraph{Iterative Distillation} Since RL training on agent tasks is time-consuming, we adopt a self-distillation approach to iteratively enhance the performance of the SFT cold-start model before resuming RL training on this improved model. Specifically, we first perform RL training on the initial cold-start model to boost agent performance. Once training has reached a certain step count or plateaued, we apply self-distillation by substituting the original cold-start data with responses generated by the RL-trained model, thus creating a superior SFT model. We then conduct further RL training on this enhanced model, progressively increasing training difficulty. This iterative strategy allows us to push the performance limits of RL-trained models efficiently.

\begin{figure}[t]
    \centering
    \includegraphics[width=0.6\linewidth]{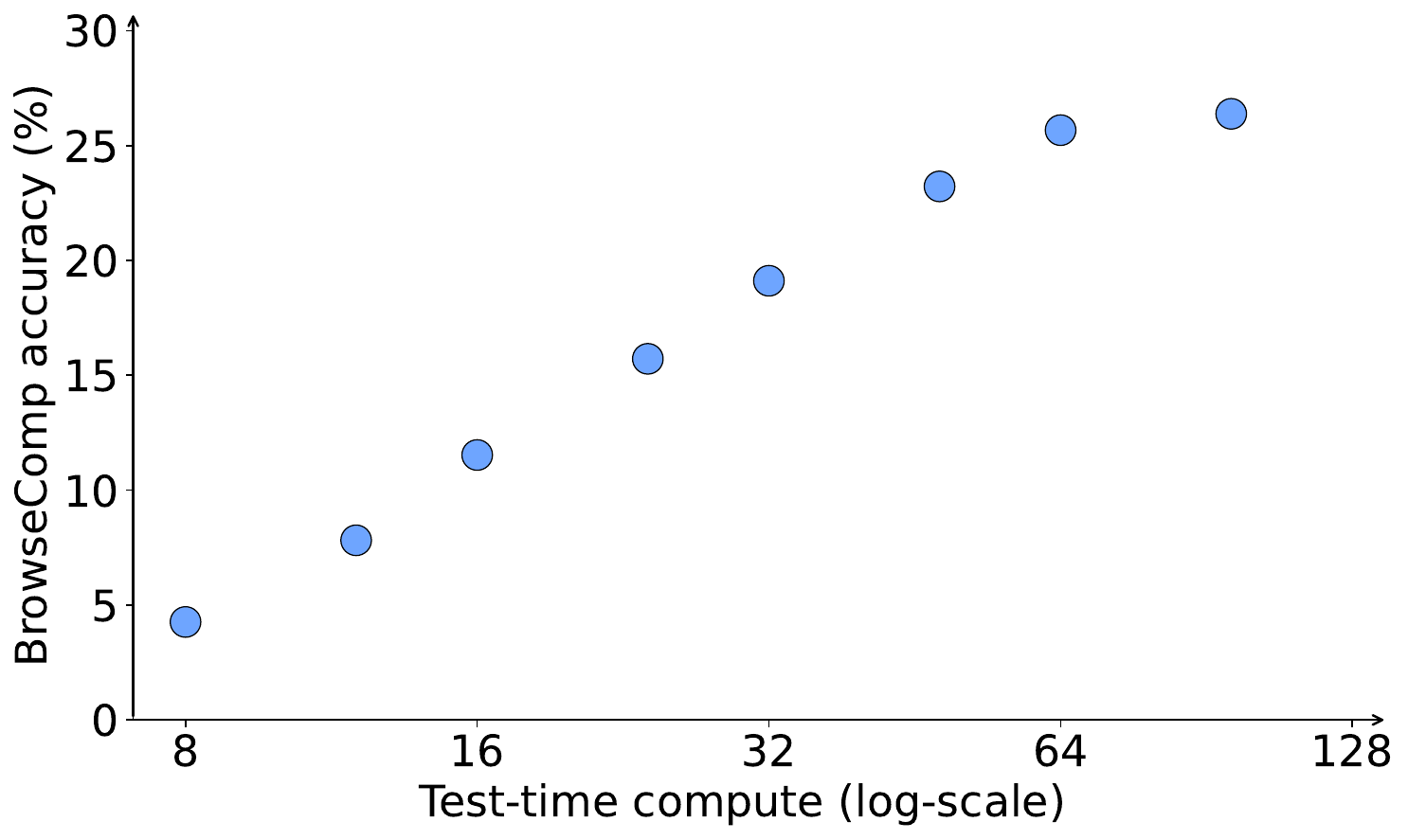}
    \caption{Interaction Turns Scaling for BrowseComp. }
    \label{fig:turn_scaling}
\end{figure}

\paragraph{Scaling Test-time Compute via Interaction Turns} For agent tasks, we observe significant performance gains given increasing interaction turns with the environment. Compared to test-time scaling in reasoning models, which scales output tokens, agent tasks make use of test-time compute by continuously interacting with the environment, e.g., searching high and low for hard-to-find web information or writing test cases for self-verification and self-correction for coding tasks. 
Figure~\ref{fig:turn_scaling} shows that with varying browsing effort,  accuracy scales smoothly with test-time compute.

\subsection{General RL}

General RL aims to holistically improve the model's overall performance, remediate potential issues, and strengthen key capabilities. Central to our methodology is a multi-source feedback system that synergizes rule-based feedback, human feedback (RLHF), and model-based feedback (RLAIF). This hybrid framework provides more robust training signals and allows us to leverage the unique advantages of each source: the precision of automated rules, the nuanced judgment of human annotators, and the scalability of AI-driven evaluation.

\paragraph{Holistic RL}

Holistic RL targets broad performance gains across diverse domains. To this end, we first construct a balanced dataset of roughly 5,000 prompts spanning 7 primary, 33 secondary, and 139 tertiary categories.
Reward signals for Holistic RL are derived from both human and AI feedback. For human feedback, we train a reward model on preference annotations. 
Annotators compare model responses and assign preference labels based on a comprehensive evaluation of multiple dimensions, such as instruction following, safety, and factual correctness.
For model feedback, we design separate scoring rubrics that depend on whether the prompt has an objective ground-truth answer.
Merging the two feedback sources yields more reliable and expressive reward signals, mitigating the inherent limitations of each individual method.

\paragraph{Instruction Following RL}

\begin{figure}[t]
    \centering
    \includegraphics[width=0.65\linewidth]{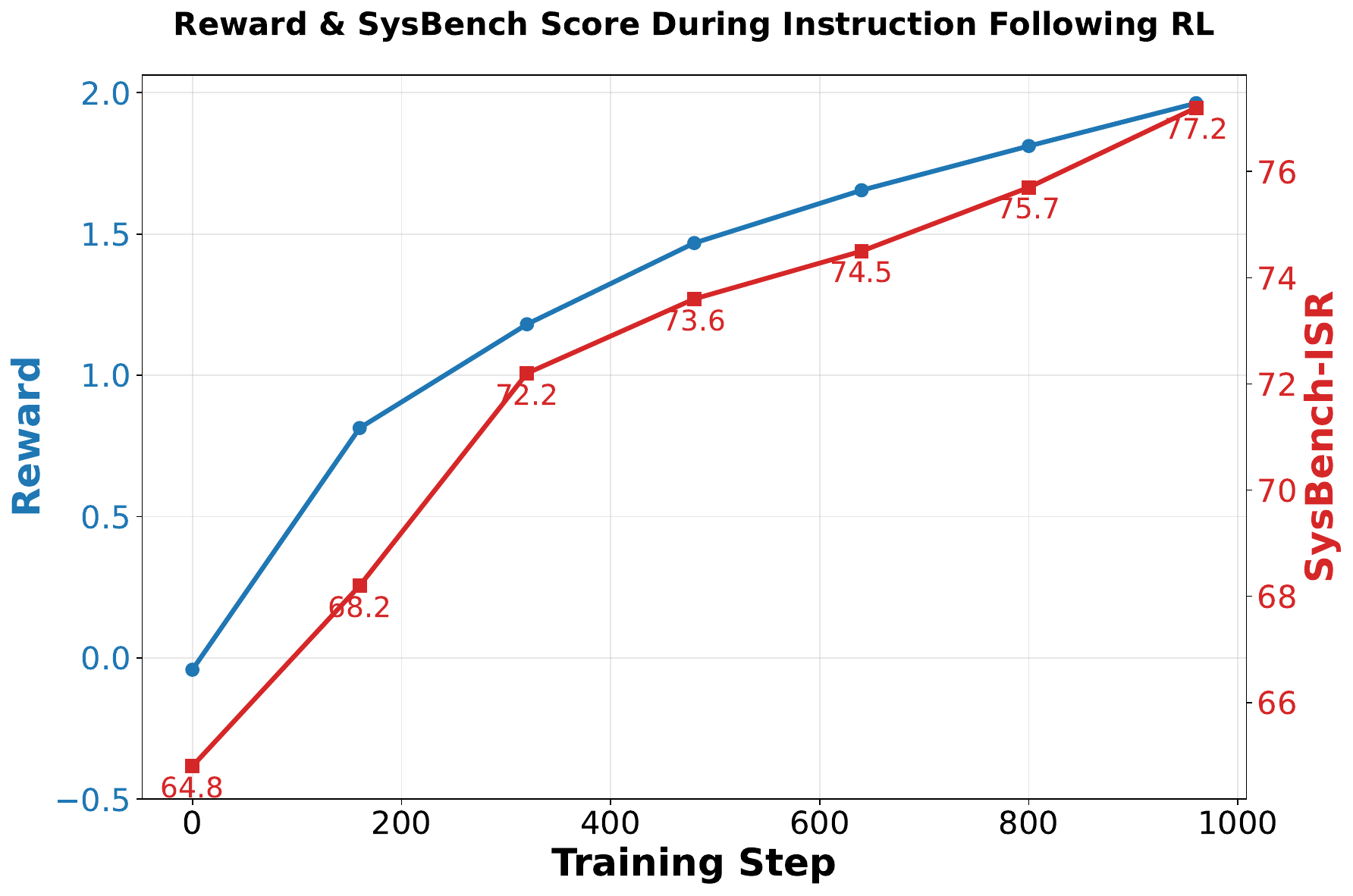}
    \caption{\textbf{Training curve of Instruction Following RL without other General RL tasks.} During GRPO training, the instruction following performance (\textcolor{red}{SysBench-ISR}) improves in step with the increasing \textcolor{blue}{reward}. Up to roughly 1,000 training steps, we have not observed clear evidence of reward hacking.}
    \label{fig:instruction-following-RL}
\end{figure}

Instruction Following RL improves the model’s ability to understand and satisfy complex instructions. To achieve this, we create a fine-grained taxonomy with 7 major and 151 minor constraint types, covering content requirements, formatting rules, and more. Based on this taxonomy, a dedicated training set of challenging instructions is assembled to cover every constraint type.
The feedback system consists of deterministic verification rules, a trained reward model, and a critique model. The robustness of this hybrid feedback system proves crucial during GRPO training. We observe mitigated reward hacking, enabling the policy model to achieve continuous and steady improvements in instruction following as shown in Figure~\ref{fig:instruction-following-RL}.

\paragraph{Function Calling RL}
Function Calling RL is divided into step-wise rule-based RL and end-to-end multi-turn RL. We incorporate step-wise rule-based RL directly into our general RL framework due to their similar output lengths and convergence speeds. For end-to-end multi-turn RL, we first train specialized expert models and then distill these experts into the main model.

\begin{itemize}[leftmargin=*,itemsep=0pt,topsep=0pt]
    \item \textbf{Step-wise Rule-based RL}: For tasks with clear tool invocation procedures, we annotate the ground truth function call for each step/turn in the training data. Given the task and the function calls from previous steps/turns, the model is trained to generate the next assistant response, which can be a function call or a response to the user. Using rule-based rewards, we guide the model to make correct function calls over consecutive rounds.
    Accordingly, we design the following strict reward function:
    $$\text{Reward} =
    \begin{cases}
    1, & \text{if } \texttt{FormatCorrect}(a_t)\ \text{and}\ \texttt{Match}(a_t, a_t^*) \\
    0, & \text{otherwise}
    \end{cases}$$
    Here, $a_t$ denotes the t-th function call generated by the model, and $a_t^*$ is the corresponding ground truth function call. A reward of 1 is only given if $a_t$ is in the correct format and matches the ground truth exactly (including the name, parameters, and every field). Otherwise, the reward is 0.
    Such a strict reward rule not only guides the model to generate correct function calls but also strongly enforces output formatting, improving the model’s usability and robustness in real-world interactions.
    \item \textbf{End-to-end Multi-turn RL}: Step-wise rule-based RL decomposes tasks into static, predetermined decision flows. In this process, the model lacks dynamic interactions with the environment and cannot autonomously explore, plan, or handle complex situations, thereby making its real-world problem-solving ability limited. To address these issues, we introduce end-to-end multi-turn function calling RL, where the model first generates the complete trajectory and is then rewarded based on task completion. In this way, the model can optimize its action policy through continuous trial and error with tool feedback, significantly enhancing its ability in autonomous planning and decision-making.
    Specifically, end-to-end multi-turn function calling RL considers two types of complex tasks: 1. single-turn multi-step tasks: The model needs to make multi-step function calls and interact with the environment to complete such tasks. We use complex tasks automatically synthesized based on MCP servers, as well as some open-source agentic datasets with runnable environments, such as Agentgym~\cite{xi2024agentgym}. 2. multi-turn multi-step tasks: Besides interacting with the tool execution environment, the model also needs to interact with an LLM-simulated user agent to obtain complete task information and accomplish the overall task.
    The reward for end-to-end multi-turn function calling RL is computed as:
    $$\text{Reward} =
    \begin{cases}
    1, & \text{if } \texttt{FormatCorrect}(a_1, \dots, a_T)\ \text{and}\ \texttt{TaskCompleted}(I, o_0,a_1, o_1, \dots, a_T,o_T) \\
    0, & \text{otherwise}
    \end{cases}$$
    Here, $I$ refers to the original complex task, $a_t$ is the t-th function call, and $o_t$ is the tool feedback or user information. $\texttt{TaskCompleted}(I, o_0,a_1, o_1, \dots, a_T,o_T)$ indicates whether the task is completed, which is determined by the environment according to predefined rules or by an LLM Judge Agent.
\end{itemize}

\paragraph{Pathology RL}

As the final stage of post-training, general RL needs to rectify potential issues, such as language mixing, excessive repetition, and formatting mistakes. Although penalizing such behaviors in the above-mentioned general RL tasks is effective, the low incidence rate of these pathologies (often less than 1\% of outputs) makes this a sample-inefficient optimization strategy. Therefore, we curate a targeted dataset for pathology RL by identifying prompts that are highly likely to trigger these pathological behaviors.
Training on this dataset lets us impose efficient penalties, further lowering the residual error rates for these problematic behaviors.

\subsection{RL Infrastructure}

\begin{figure}[t]
    \centering
    \includegraphics[width=0.9\linewidth, trim=0 10 0 10, clip]{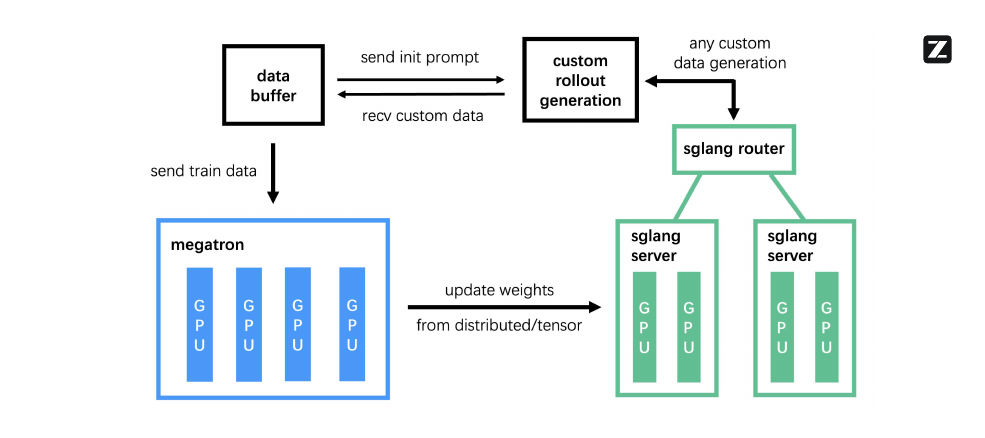}
    \caption{Overview of the Slime RL infrastructure. The system consists of three core modules: \textbf{Training (Megatron)} – handles the main training process, reads data from the Data Buffer, and synchronizes parameters with the rollout module after training; \textbf{Rollout (SGLang + Router)} – generates new data, including rewards and verifier outputs, and writes it to the Data Buffer; \textbf{Data Buffer} – serves as a bridge module that manages prompt initialization, custom data, and rollout generation strategies.}
    \label{fig:slime}
\end{figure}

Our RL infrastructure is built upon Slime\footnote{\url{https://github.com/THUDM/slime}}, an open-source framework we developed. The framework is engineered with several key optimizations to enhance flexibility, efficiency, and extensibility.

\paragraph{Flexible Hybrid Training and Data Generation Architecture} A core feature of our infrastructure is its support for highly flexible training paradigms and data generation strategies within a single, unified system. This design allows us to cater to the distinct requirements of various RL tasks by supporting both a colocated, synchronous mode and a disaggregated, asynchronous mode. This flexibility in data generation is crucial for extending our RL capabilities to new domains and more complex agentic environments.
We observed that different RL tasks benefit from different scheduling approaches. For general-purpose RL tasks or those aimed at enhancing model reasoning capabilities (e.g., in mathematics and code generation), a synchronous, colocated architecture is more effective. In this setup, the training and inference engines reside on the same worker. This, combined with dynamic sampling, significantly reduces GPU idle time and maximizes resource utilization.
Conversely, for agentic tasks, such as those in Software Engineering (SWE), the data generation process is often protracted and involves complex system interactions. To ensure that the agent environments can operate continuously and maximize data throughput, we adopt a disaggregated, asynchronous model. The rollout component of the RL framework is exposed directly to the agent environment, while the GPUs for training and inference are scheduled independently. This decoupling enables the agent environments to constantly generate new data without being stalled by the training cycle.
By leveraging the resource scheduling and asynchronous capabilities of the Ray framework, we can flexibly place the inference and training engines on the same GPU or on different ones. This dual support for synchronous and asynchronous training allows diverse RL tasks to share a common set of underlying optimizations for both training and inference.

\paragraph{Accelerated Rollout with Mixed-Precision Inference} Rollout efficiency is a persistent bottleneck in RL training. To address this, our infrastructure supports BF16 for training while leveraging FP8 for inference to accelerate the data generation phase.
During each policy update iteration, we perform online, block-wise FP8 quantization on the model parameters before they are dispatched for rollout. This dynamic quantization enables highly efficient FP8 inference, significantly improving the overall throughput of the data collection process.

\paragraph{Agent-oriented RL Infra Design}

To conduct RL for agent tasks, we design a fully asynchronous and decoupled RL infrastructure that efficiently handles long-horizon agent rollouts and supports flexible multi-task RL training across diverse agent frameworks.

Agentic rollouts often require prolonged interactions with complex environments, which can significantly slow down the overall RL training process. To overcome this, we first design a high-concurrency Docker-based runtime that provisions isolated environments for each task, drastically reducing rollout overhead. In addition, we implement a fully asynchronous RL training loop. Because agent tasks can vary in type and trajectory length, synchronous RL training often leads to severe GPU underutilization as workers wait for the slowest rollouts to complete. Our approach partitions GPUs into dedicated rollout engines and training engines: the rollout engines continuously generate trajectories, while the training engines update the model weights and periodically synchronize them back to the rollout engines. This decoupled design prevents long or diverse trajectories from blocking the entire training pipeline, resulting in consistently high throughput, particularly in scenarios with highly variable agent interactions.

Another key challenge is the diversity of existing agent frameworks, which are tailored to different tasks. Leveraging these frameworks not only improves task-specific performance but also maintains alignment between training and inference. To achieve this, we introduce a unified HTTP endpoint interface coupled with a centralized data pool. Since most agent frameworks produce rollouts in a message-list format, all trajectories are stored in this data pool, which serves as a shared source for training. This architecture cleanly decouples task-specific rollout logic from the RL training process, enabling seamless integration of heterogeneous agent frameworks. Furthermore, the data pool supports customizable, task-specific filtering and dynamic sampling strategies to ensure high-quality RL training data across diverse tasks.

Through these two core designs, our system provides a scalable, flexible, and high-performance solution for long-agentic RL, and can support long-horizon rollouts and adapt to a wide range of agent tasks.

\section{Evaluation}
\label{sec:evaluation}
\subsection{Evaluation of Base Models}
We first evaluate the performance of our base model GLM-4.5-Base. \Cref{tab:base_benchmark} shows the comparison results of the last checkpoint of pre-training of our base model. Please note that the base model has not been trained on instruction data, and the GLM-4.5-Base scores are from our internal evaluation framework.
Results show that GLM-4.5-Base is stable on all the different benchmarks, including English, Code, Math, and Chinese, which validates our idea of unifying all the abilities into one model.

\begin{table}[htbp]
  \centering
  \caption{Comparison among GLM-4.5-Base and other representative open-source base models.}
  \resizebox{\textwidth}{!}{
  \begin{tabular}{lccccc|c}
    \toprule
     & \textbf{Benchmark (Metric)} & \textbf{Qwen3-235B} & \textbf{Llama4-Maverick} & \textbf{DeepSeek-V3} & \textbf{Kimi-K2} & \textbf{GLM-4.5} \\
    & & \textbf{-A22B Base} & \textbf{400B Base} & \textbf{Base} & \textbf{Base} & \textbf{Base} \\
    \midrule
    \multirow{3}{*}{} & Architecture & MoE & MoE & MoE & MoE & MoE \\
    & \# Activated Params & 22B & 17B & 37B & 32B & 32B \\
    & \# Total Params & 235B & 400B & 671B & 1043B & 355B \\
    \midrule
    \multirow{6}{*}{\textbf{English}} & SimpleQA (EM) & - & - & 26.6 & 35.3 & 30.0 \\
    & BBH (EM) & 88.9 & 87.1 & 88.4 & 88.7 & 86.2 \\
    & MMLU (EM) & 87.8 & 85.2 & 87.2 & 87.8 & 86.1 \\
    & HellaSwag (EM) & - & - & 88.9 & 94.6 & 87.1 \\
    & PIQA (EM) & - & - & 84.7 & - & 85.3 \\
    & TriviaQA (EM) & - & - & 82.9 & 85.1 & 80.0 \\
    \midrule
    \multirow{3}{*}{\textbf{Code}} & EvalPlus (Pass@1) & 77.6 & 65.5 & 65.6 & 80.3 & 78.1 \\
    & LiveCodeBench-Base (Pass@1) & - & 25.1 & 24.6 & 26.3 & 28.1 \\
    \midrule
    \multirow{2}{*}{\textbf{Math}} & GSM8K (EM) & 94.4 & 87.7 & 87.6 & 92.1 & 79.4\\
    & MATH (EM) & 71.8 & 63.3 & 62.6 & 70.2 & 61.0 \\
    \midrule
    \multirow{3}{*}{\textbf{Chinese}} & CLUEWSC (EM) & - & - & 82.7 & - & 83.5 \\
    & C-Eval (EM) & - & 80.9 & 90.1 & 92.5 & 86.9 \\
    & C3 (EM) & - & - & 78.6 & - & 83.1 \\
    & Chinese-SimpleQA (EM) & - & 53.5 & 72.1 & 77.6 & 70.1 \\
    \bottomrule
  \end{tabular}
  }
  \label{tab:base_benchmark}
\end{table}

\subsection{Evaluation on 12 (ARC) Benchmarks}

We further evaluate our full GLM-4.5 models after Post-Training for all the Agentic, Reasoning, and Coding (ARC) tasks, on 12 benchmarks: MMLU-Pro, AIME 24, MATH-500, SciCode, GPQA, HLE, LCB (2407-2501), SWE-Bench Verified, Terminal-Bench, TAU-Bench, BFCL V3, BrowseComp.

\subsubsection{Evaluation of Agentic Abilities}
\begin{table}[ht]
    \centering
    \footnotesize
    \caption{Results on Agentic Benchmarks. TAU represents TAU-bench~\cite{yao2024tau} and BFCL represents Berkeley Function Calling Leaderboard~\cite{patil2025bfcl}.}
    \begin{tabular}{@{}l@{\;\;\;}c@{\;\;\;}c@{\;\;\;}c@{\;\;\;}c@{\;\;\;}c@{\;\;\;}c@{\;\;\;}c@{\;\;\;}c@{\;\;\;}c@{\;\;\;}c@{\;\;\;}c@{\;\;\;}c@{}}
    \toprule
    Benchmark & \makecell[c]{\textbf{GLM-}\\\textbf{4.5}} & \makecell[c]{\textbf{GLM-}\\\textbf{4.5-Air}} & \makecell[c]{\textbf{o3}} & \makecell[c]{\textbf{o4}\\\textbf{mini}} & \textbf{GPT-4.1} & \makecell[c]{\textbf{Claude}\\\textbf{Opus 4}} & \makecell[c]{\textbf{Claude}\\\textbf{Sonnet 4}} & \makecell[c]{\textbf{Gemini}\\\textbf{2.5 Pro}} & \makecell[c]{\textbf{Kimi}\\\textbf{K2}} & \textbf{Grok 4} \\
    \midrule
    TAU-Retail       & 79.7 & 77.9 & 70.4 & 65.6 & 75.1 & 81.4 & 80.5 & 77.0 & 73.9 & 76.5 \\
    TAU-Airline      & 60.4 & 60.8 & 52.0 & 49.2 & 48.8 & 59.6 & 60.0 & 48.0 & 51.2 & 58.4 \\
    BFCL V3  & 77.8 & 76.4 & 72.4 & 67.2 & 68.9 & 74.4 & 75.2 & 61.2 & 71.1 & 66.2 \\
    BrowseComp       & 26.4 & 21.3 & 49.7 & 28.3 & 4.1  & 18.8 & 14.7 & 7.6   & 7.9  & 32.6 \\
    \midrule
    Average & 58.1 & 55.7 & 61.1 & 50.1 & 45.0 & 54.6 & 53.4 & 43.8 & 47.2 & 55.4\\
    \bottomrule
    \end{tabular}
    \label{tab:agentic_coding}
\end{table}

We evaluate the agentic abilities of GLM-4.5 in two aspects: TAU-bench~\cite{yao2024tau} (including retail and airline domains) and Berkeley Function Call Leaderboard V3 (BFCL V3)~\cite{patil2025bfcl}, which measures the model's ability to call user-defined functions to respond to users' queries. BrowseComp~\cite{wei2025browsecomp} measures the model's ability as a web browsing agent to find correct answers for complicated questions. For TAU-bench, we use an optimized user simulator (Cf.~\Cref{fig:tau-prompt}) for both the Retail and Airline domains. The user prompt we use can be found below in~\Cref{fig:tau-prompt}. 
On TAU-bench, GLM-4.5's performance is better than Gemini 2.5 Pro and close to Claude Sonnet 4. On BFCL V3, GLM-4.5 achieves the best overall score among the baselines. On BrowseComp, the performance of OpenAI o3 is much better than that of other models. GLM-4.5's performance is close to the second-best model (o4-mini) and significantly better than Claude Opus 4.

\begin{tcolorbox}[left=0mm,right=0mm,top=0mm,bottom=0mm,boxsep=1mm,arc=0mm,boxrule=0pt, frame empty, breakable]
\footnotesize
\begin{lstlisting}
You are a user interacting with an agent.{instruction_display}
# Rules:
- Just generate one line at a time to simulate the user's message.
- Do not give away all the instruction at once. Only provide the information that is necessary for the current step.
- Do not hallucinate information that is not provided in the instruction. Follow these guidelines:
 1. If the agent asks for information NOT in the instruction:
 - Say you don't remember or don't have it
 - Offer alternative information that IS mentioned in the instruction
 2. Examples:
 - If asked for order ID (not in instruction): ``Sorry, I don't remember the order ID, can you search for it? My name/email/phone number/zipcode is ...''
 - If asked for email (not in instruction): ``I don't have my email handy, but I can give you my name and zip code which are...''
- Do not repeat the exact instruction in the conversation. Instead, use your own words to convey the same information.
- Try to make the conversation as natural as possible, and stick to the personalities in the instruction.
# Constraint Handling:
- Provide requests strictly based on what is explicitly stated in the instruction.
- Do not assume, extend, substitute, or generalize in any form.
- Do not modify or relax constraints on:
- Time / Date
- Budget
- Specific terms (e.g., ``same'' must not be replaced with ``similar'')
- Core Rule: Any attribute NOT mentioned in the instruction can be either changed or kept the same
- Examples:
 - If instruction says ``exchange red item to blue'': Only color must change, other attributes (size, material, etc.) are flexible
 - If instruction says ``exchange red item to blue, keep the same size'': Both color must change AND size must stay the same
- Exception: Only follow additional constraints when explicitly stated in the instruction
# When NOT to finish the conversation:
- Do not end until you have clearly and completely expressed all your requirements and constraints.
- Do not end until the agent has completed all tasks mentioned in the instruction and verified no operations were missed.
- Do not end if the agent's execution results do not match your expectations or are incorrect/incomplete.
# When you CAN finish the conversation:
- Only when all above conditions are satisfied AND all tasks are completed correctly.
- OR when you have clearly expressed complete requirements but the system explicitly states it cannot complete them due to technical limitations - in this case, accept transfer to human.
# How to finish the conversation:
- If the agent has completed all tasks, generate ``###STOP###'' as a standalone message without anything else to end the conversation.
# Note:
- You should carefully check if the agent has completed all tasks mentioned in the instruction before generating ``###STOP###''.
\end{lstlisting}
\end{tcolorbox}
\noindent\begin{minipage}{\textwidth}
\captionof{figure}{One example of user prompt we used for TAU-bench.}
\label{fig:tau-prompt}
\end{minipage}

\subsubsection{Evaluation of Reasoning}

We evaluate the reasoning abilities of GLM-4.5 and GLM-4.5-Air on seven benchmarks, including MMLU-Pro~\cite{wang2024mmlu}, AIME 24, MATH 500~\cite{hendrycks2measuring}, SciCode~\cite{tian2024scicode}, GPQA~\cite{rein2024gpqa}, Humanity's Last Exam (HLE)~\cite{phan2025humanity}, and LiveCodeBench (LCB)~\cite{jainlivecodebench}\footnote{LiveCodeBench is a dynamic benchmark and we evaluate on problems between 7/1/2024 and 1/1/2025.}. For the AIME and GPQA benchmarks, we report the average accuracy over 32 and 8 samples, respectively (Avg@32, Avg@8), to mitigate result variance. An LLM was used for automated answer validation. For the HLE benchmark, only the text-based questions were evaluated, with correctness judged by GPT-4o. Our evaluation code is also open-sourced\footnote{\url{https://github.com/zai-org/glm-simple-evals}}. We also compute the average reasoning performance on the seven benchmarks with the intelligence index proposed by Artificial Analysis\footnote{\url{https://artificialanalysis.ai}}. GLM-4.5 outperforms OpenAI o3 on AIME 24 and SciCode. On average, GLM-4.5 outperforms Claude Opus 4 and is close to DeepSeek-R1-0528.

\begin{table}[ht]
    \centering
    \footnotesize
    \caption{Results on Reasoning Benchmarks. HLE represents Humanity's Last Exam~\cite{phan2025humanity} and LCB represents LiveCodeBench (2407-2501)~\cite{jainlivecodebench}.}
    \begin{tabular}{lcccccccccccc}
    \toprule
    Benchmark&\makecell[c]{\textbf{GLM-}\\\textbf{4.5}} & \makecell[c]{\textbf{GLM-}\\\textbf{4.5-Air}} & \makecell[c]{\textbf{o3}} & \makecell[c]{\textbf{Claude}\\\textbf{Opus 4}} & \makecell[c]{\textbf{Gemini}\\\textbf{2.5 Pro}} & \makecell[c]{\textbf{DeepSeek}\\\textbf{R1 0528}} & \makecell[c]{\textbf{Qwen3}\\\textbf{235B 2507}} & \textbf{Grok 4} \\
    \midrule
    MMLU Pro         & 84.6 & 81.4 & 85.3 & 87.3 & 86.2 & 84.9 & 84.5 & 86.6 \\
    AIME 24          & 91.0 & 89.4 & 90.3 & 75.7 & 88.7 & 89.3 & 94.1 & 94.3 \\
    MATH 500         & 98.2 & 98.1 & 99.2 & 98.2 & 96.7 & 98.3 & 98.0 & 99.0 \\
    SciCode          & 41.7 & 37.3 & 41.0 & 39.8 & 42.8 & 40.3 & 42.9 & 45.7 \\
    GPQA             & 79.1 & 75.0 & 82.7 & 79.6 & 84.4 & 81.3 & 81.1 & 87.7 \\
    HLE              & 14.4 & 10.6 & 20.0 & 11.7 & 21.1 & 14.9 & 15.8 & 23.9 \\
    LCB  & 72.9 & 70.7 & 78.4 & 63.6 & 80.1 & 77.0 & 78.2 & 81.9 \\
    \midrule
    AA-Index (Est.)  & 67.7 & 64.8 & 70.0 & 64.4 & 70.5 & 68.3 & 69.4 & 73.2 \\
    \bottomrule
    \end{tabular}
    \label{tab:reasoning}
\end{table}

\subsubsection{Evaluation of Coding}

\begin{table}[ht]
    \centering
    \footnotesize
    \caption{Results on SWE-bench Verified and Terminal-Bench}
    \begin{tabular}{@{}l@{\;\;}c@{\;\;\;}c@{\;\;\;}c@{\;\;}c@{\;\;\;}c@{\;\;\;}c@{\;\;\;}c@{\;\;\;}c@{\;\;\;}c@{\;\;\;}c@{\;\;\;}c@{\;\;\;}c@{\;}c@{}}
    \toprule
    Benchmark & \makecell[c]{\textbf{GLM-}\\\textbf{4.5}} & \makecell[c]{\textbf{GLM-}\\\textbf{4.5-Air}} & \textbf{o3} & \textbf{GPT-4.1} & \makecell[c]{\textbf{Claude}\\\textbf{Opus 4}} & \makecell[c]{\textbf{Claude}\\\textbf{Sonnet 4}} & \makecell[c]{\textbf{Gemini}\\\textbf{2.5 Pro}} & \makecell[c]{\textbf{DeepSeek}\\\textbf{R1 0528}} & \textbf{Kimi K2} \\
    \midrule
    SWE-bench Verified & 64.2 & 57.6 & 69.1 & 48.6 & 67.8 & 70.4 & 49.0 & 41.4 & 65.4 \\
    Terminal-Bench     & 37.5 & 30.0 & 30.2 & 30.3 & 43.2 & 35.5 & 25.3 & 17.5 & 25.0 \\
    \midrule
    Average & 50.9 & 43.8 & 49.7 & 39.5 & 55.5 & 53.0 & 37.2 & 29.5 & 45.2 \\
    \bottomrule
    \end{tabular}
    \label{tab:swe_terminal}
\end{table}

To measure GLM-4.5's ability to complete real-world coding tasks, we evaluate it on two challenging benchmarks, SWE-bench Verified~\cite{jimenez2023swe} and Terminal-Bench~\cite{tbench_2025}. SWE-bench measures the model's ability to modify an existing codebase to solve a GitHub issue. The Verified subset is a human-filtered subset of 500 instances. For evaluation, we use OpenHands~\cite{wang2025openhands} v0.34.0 with runs limited to 100 iterations and history truncation to prevent exceeding the 128K context limit, configured with temperature=0.6, top\_p=1.0. Terminal-Bench measures the model's ability to accomplish complex tasks in a terminal environment. We use the Terminus framework and standard function calling rather than direct prompting for evaluation. On SWE-bench Verified, GLM-4.5 outperforms GPT-4.1 and Gemini-2.5-Pro. On Terminal-Bench, GLM-4.5 outperforms Claude Sonnet 4. On average, GLM-4.5 is the best competitor for Claude Sonnet 4 on coding tasks.

\subsubsection{Evaluation of General Abilities}

\begin{table}[!ht]
    \centering
    \scriptsize
    \setlength{\tabcolsep}{3pt}
    \caption{Results on Commonly Used General Chat Benchmarks}
    \begin{tabularx}{\linewidth}{l>{\centering\arraybackslash}X>{\centering\arraybackslash}X>{\centering\arraybackslash}X>{\centering\arraybackslash}X>{\centering\arraybackslash}p{0.8cm}>{\centering\arraybackslash}X>{\centering\arraybackslash}X>{\centering\arraybackslash}X>{\centering\arraybackslash}X>{\centering\arraybackslash}X}
    \toprule
        Benchmark & \makecell[c]{\textbf{GLM-}\\\textbf{4.5}} & \makecell[c]{\textbf{GLM-}\\\textbf{4.5-Air}} & \makecell[c]{\textbf{GPT-4.1}} & \makecell[c]{\textbf{Claude}\\\textbf{Sonnet 4}} & \makecell[c]{\textbf{Gemini}\\\textbf{2.5 Pro}} & \textbf{Grok 4} & \makecell[c]{\textbf{Qwen3}\\\textbf{235B}} & \makecell[c]{\textbf{Deepseek}\\\textbf{R1 0528}} & \makecell[c]{\textbf{DeepSeek}\\\textbf{V3 0324}} & \textbf{Kimi K2} \\ 
    \midrule
        MMLU & 90.0  & 87.4  & 90.2  & 91.9  & 91.9  & 91.9  & 90.2  & 89.9  & 89.1  & 89.5  \\ 
        SimpleQA & 26.4  & 14.5  & 42.3  & 18.5  & 54.0  & 51.9  & 45.8  & 27.8  & 27.7  & 31.0   \\ 
        IFEval & 86.1  & 86.3  & 87.4  & 88.7  & 90.8  & 92.4  & 87.8  & 80.0  & 83.4  & 89.8  \\ 
        SysBench & 81.0  & 77.4  & 80.6  & 80.6  & 82.2  & 81.5  & 83.3  & 81.2  & 79.8  & 79.0  \\ 
        MultiChallenge & 52.8  & 42.5  & 38.3  & 55.3  & 57.5  & 65.2  & 58.2  & 46.5  & 37.0  & 54.1  \\
    \bottomrule
    \end{tabularx}
    \label{tab:general_chat}
\end{table}

To evaluate the model's general abilities, we employed a set of widely-adopted open-source benchmark datasets, encompassing knowledge-intensive evaluations MMLU~(EM)~\citep{hendrycks2measuring} and SimpleQA~(Correct)~\citep{simpleqa}, and instruction-following assessments IFEval~(Prompt Strict)~\citep{zhou2023instruction}, SysBench~(ISR)~\citep{qinsysbench}, and MultiChallenge~\citep{deshpande2025multichallenge}. MultiChallenge is a multi-turn conversational benchmark evaluating LLMs across four integrated capability dimensions. SysBench systematically evaluates LLMs' system message following capabilities across multi-turn conversations through three-level granularity metrics. On the MMLU benchmark, nearly all flagship models, including GLM-4.5, demonstrate performance at a comparable level. SimpleQA, which reflects the factual knowledge of a model, shows that GLM-4.5 (355B) performs similarly to DeepSeek V3 and R1 (both 671B), despite having nearly half of the parameters. On the IFEval benchmark, GLM-4.5 outperforms DeepSeek R1. In the Sysbench evaluation, GLM-4.5 surpasses GPT-4.1, DeepSeek V3, and Kimi K2. Additionally, on the MultiChallenge benchmark, it demonstrates superior performance compared to both GPT-4.1 and DeepSeek R1.

\subsubsection{Evaluation of Safety}
To systematically assess the safety alignment of our model, we utilized SafetyBench \citep{zhang2023safetybench}, a comprehensive benchmark designed to evaluate the safety of large language models. SafetyBench consists of 11,435 multiple-choice questions covering seven distinct categories of safety concerns, with data in both English and Chinese. This benchmark enables a standardized and scalable evaluation of a model's ability to handle potentially harmful or sensitive topics. The categories include Ethics and Morality, Illegal Activities, Mental Health, Offensiveness, Physical Health, Privacy and Property, and Unfairness and Bias.
We evaluated GLM-4.5 against a suite of other leading models. The results indicate that GLM-4.5 achieves a strong safety score, competitive with other top-tier models. Its overall score of 89.87 is comparable to that of Kimi-K2 (90.48) and GPT-4.1 (89.71). Notably, GLM-4.5 demonstrates robust performance in the areas of Ethics and Morality (94.33), Mental Health (94.67), and Physical Health (96.67). While it performs well in preventing responses related to Illegal Activities (90.97) and protecting Privacy and Property (92.00), there is still room for improvement in addressing Unfairness and Bias, an area of ongoing focus for our development efforts.
The detailed performance breakdown is presented in the table below.

\begin{table}[h!]
\centering
\caption{Evaluation Results on SafetyBench}
\label{tab:safetybench-results}
\footnotesize
\begin{tabular}{@{}l@{}c@{\;\;}c@{\;\;}c@{\;\;}c@{\;\;}c@{\;\;}c@{\;\;}c@{\;\;}c@{}}
\toprule
Model & \textbf{Average} & \makecell{\textbf{Ethics} \& \\\textbf{Morality}} & \makecell{\textbf{Illegal}\\\textbf{Activities}} & \makecell{\textbf{Mental}\\\textbf{Health}} & \makecell{\textbf{Offensiveness}} & \makecell{\textbf{Physical}\\\textbf{Health}} & \makecell{\textbf{Privacy} \&\\\textbf{Property}} & \makecell{\textbf{Unfairness}\\\& \textbf{Bias}} \\
\midrule
{GLM-4.5} & {89.9} & {94.3} & {91.0} & {94.7} & {83.0} & {96.7} & {92.0} & {77.4} \\
GLM-4.5-Air & 87.8 & 91.0 & 90.3 & 92.7 & 83.3 & 92.3 & 90.3 & 74.7 \\
Gemini 2.5 Pro & 90.5 & 94.7 & 91.7 & 95.3 & 84.3 & 97.0 & 92.3 & 78.0 \\
Kimi K2 & 90.5 & 93.0 & 93.3 & 95.0 & 90.3 & 97.3 & 93.0 & 71.3 \\
GPT-4.1 & 89.7 & 92.0 & 94.3 & 95.3 & 85.3 & 95.7 & 91.3 & 74.0 \\
DeepSeek-V3-0324 & 88.8 & 92.3 & 90.7 & 95.0 & 84.7 & 95.7 & 91.0 & 72.3 \\
DeepSeek-R1-0528 & 83.5 & 87.0 & 81.7 & 86.7 & 77.7 & 92.0 & 85.7 & 73.7 \\
\bottomrule
\end{tabular}%
\end{table}


\subsection{Evaluations for Hands-on Experience}

Sometimes, a trained LLM may overfit some predefined benchmarks, which makes the evaluated results not precisely reflect real-world experience.
To overcome this challenge and to gauge our model's performance in more realistic situations, we have established a comprehensive manual evaluation framework. Human evaluation is particularly advantageous for assessing performance on open-ended questions, where aspects like coherence, relevance, and creativity are paramount. This hands-on approach allows for a more granular analysis, enabling us to better pinpoint areas of weakness and understand the qualitative aspects of model behavior that automated metrics often miss.

\subsubsection{Evaluation of General Chat}
To test the practical application capabilities of our models, we curated a diverse dataset of real-scenario user prompts. These prompts span multiple languages and cover a wide range of categories, including Mathematics, Text Processing, Text Generation, Subjective QA, Objective QA, Logical Reasoning, and Code Instructions. We meticulously filtered this collection to ensure high quality and appropriate difficulty, while also removing any data that could compromise user privacy or safety. The final dataset consists of 660 prompts, with a distribution of 392 in English, 108 in Chinese, and 160 in other languages. For prompts requiring factual knowledge, we annotated the correct answers to serve as a ground truth for evaluation.

We conducted a comparative evaluation between GLM-4.5, Deepseek-R1-0528, and Kimi K2. For each prompt, the responses from the different models were presented in a randomized order to eliminate any potential sequential bias. A single, consistent evaluator then scored each response on a scale of 0 to 10. This method of using the same evaluator at the same time for a batch of comparisons is designed to minimize deviations arising from different individual preferences and subjective standards. The reasoning contents of GLM-4.5 and Deepseek-R1-0528 are not presented to the evaluators.
The average scores for each model across the different categories and languages are presented below.

\paragraph{English Results}
In the English prompt set, GLM-4.5 achieved the highest overall score. It demonstrated particularly strong performance in Mathematics, Objective QA, and Text Generation.
\begin{table}[h!]
\centering
\caption{Human Evaluation Scores on English Prompts. Subj. stands for Subjective. Obj stands for Objective. Text Gen. stands for Text Generation.}
\label{tab:eng-results}
\footnotesize
\begin{tabular}{@{}l@{}cccccccc@{}}
\toprule
\textbf{Model} & \textbf{Overall} & \textbf{Math} & \textbf{Text Proc.} & \textbf{Subj. QA} & \textbf{Obj. QA} & \textbf{Text Gen.} & \textbf{Logic}& \textbf{Code} \\
\midrule
GLM-4.5 & 8.66 & 8.72 & 8.00 & 8.36 & 8.82 & 8.61 & 9.25 & 8.53 \\
DeepSeek-R1-0528 & 8.62 & 8.56 & 8.27 & 7.91 & 9.00 & 7.83 &9.07 & 8.65 \\
Kimi-K2 & 8.13 & 7.22 & 8.00 & 7.45 & 8.86 & 7.06 &7.07 & 8.71 \\
\bottomrule
\end{tabular}
\end{table}

\paragraph{Chinese Results}
For the Chinese prompts, GLM-4.5 again led with the highest average score, showing standout performance in Text Generation, Logical Reasoning, and Code Instructions.
\begin{table}[h!]
\centering
\caption{Manual Evaluation Scores on Chinese Prompts}
\label{tab:chn-results}
\footnotesize
\begin{tabular}{@{}l@{}cccccccc@{}}
\toprule
\textbf{Model} & \textbf{Overall} & \textbf{Math} & \textbf{Text Proc.} & \textbf{Subj. QA} & \textbf{Obj. QA} & \textbf{Text Gen.} & \textbf{Logic} & \textbf{Code} \\
\midrule
GLM-4.5 & 8.37 & 7.68 & 8.20 & 8.50 & 8.66 & 9.00 & 9.27 & 8.89 \\
DeepSeek-R1-0528 & 8.05 & 7.76 & 8.07 & 8.00 & 7.89 & 8.59 & 9.00 & 8.67 \\
Kimi-K2 & 7.03 & 7.37 & 6.43 & 7.71 & 6.45 & 8.28 & 7.55 & 8.26 \\
\bottomrule
\end{tabular}
\end{table}

\paragraph{Other Languages Results}
In the multilingual evaluation covering other languages, GLM-4.5 maintained its lead, excelling in Text Generation and Subjective QA.
\begin{table}[h!]
\centering
\caption{Manual Evaluation Scores on Other Language Prompts}
\label{tab:other-lang-results}
\footnotesize
\begin{tabular}{@{}l@{}cccccccc@{}}
\toprule
\textbf{Model} & \textbf{Overall} & \textbf{Math} & \textbf{Text Proc.} & \textbf{Text Gen.} & \textbf{Subj. QA} & \textbf{Obj. QA} & \textbf{Code} & \textbf{Logic} \\
\midrule
GLM-4.5 & 8.49 & 8.67 & 8.13 & 8.90 & 9.33 & 8.71 & 7.86 & 8.33 \\
DeepSeek-R1-0528 & 8.27 & 9.44 & 8.38 & 7.86 & 9.44 & 8.22 & 7.64 & 8.17 \\
Kimi-K2 & 6.63 & 7.22 & 6.38 & 7.62 & 7.78 & 6.22 & 6.68 & 7.17 \\
\bottomrule
\end{tabular}
\end{table}

\subsubsection{Evaluation of Coding Agent}

\begin{figure}[ht]
   \centering
   \includegraphics[width=1.0\linewidth]{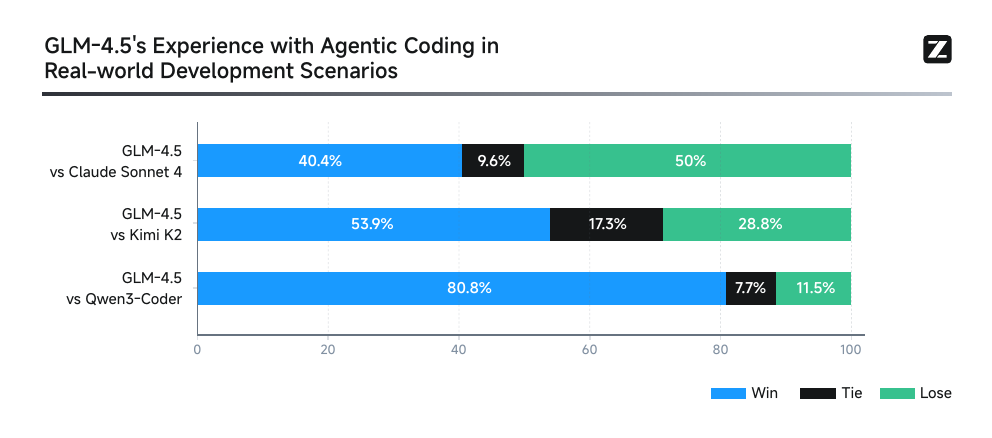}
  \caption{Head-to-head evaluation results between GLM-4.5 and other models on CC-Bench.}
  \label{fig:winrate}
\end{figure}

\begin{figure}[ht]
   \centering
   \includegraphics[width=1.0\linewidth]{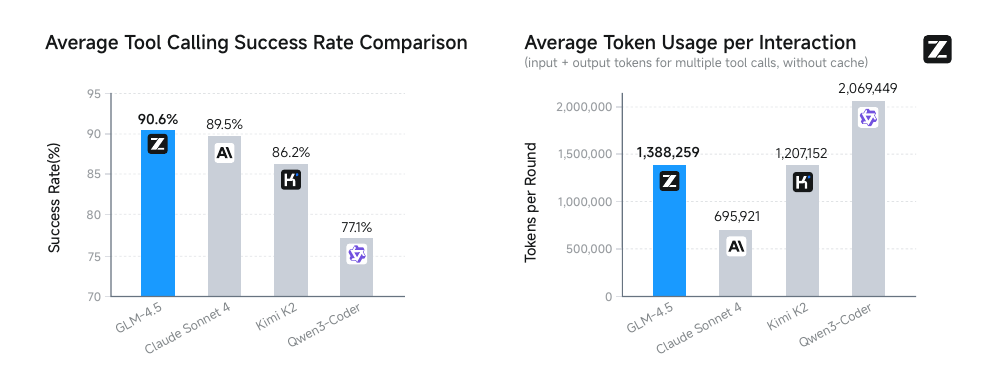}
  \caption{Average tool calling success rate and token usage per interaction across different models on CC-Bench.}
  \label{fig:successrate}
\end{figure}

\paragraph{Experimental Setup} To evaluate the agentic coding capabilities of GLM-4.5 in real-world scenarios, we constructed \textbf{CC-Bench}, a benchmark built on the Claude Code\footnote{\url{https://github.com/anthropics/claude-code}}, encompassing 52 carefully designed programming tasks across diverse software development domains\footnote{The detailed task descriptions and all evaluation trajectories of CC-Bench are available at \url{https://huggingface.co/datasets/zai-org/CC-Bench-trajectories}}. We compare GLM-4.5 against three strong baselines: Claude Sonnet 4, Kimi K2, and Qwen3-Coder.
Each task was executed in an isolated containerized environment to prevent cross-task interference, with models initialized using predefined API configurations. Testing was conducted interactively by human experts over multiple rounds: each task began with a standardized prompt, followed by iterative interactions where experts adjusted inputs based on model outputs until the task was completed or failed. To ensure fairness, the same expert followed consistent interaction strategies across all models.

Based on this testing procedure, model performance was evaluated using the following criteria: the primary metric was \textbf{task completion}, determined by predefined completion criteria. In cases of ties, \textbf{efficiency and reliability}—including tool calling success rate and token consumption efficiency—were used as secondary metrics. The evaluation prioritized functional correctness and task completion over efficiency metrics, ensuring that coding capability remained the primary evaluation focus.

\paragraph{Results} In head-to-head evaluations, GLM-4.5 demonstrated strong performance relative to open-source baselines and competitive capability against closed-source models as shown in figure~\ref{fig:winrate}. Specifically:

\begin{itemize}
  \item \textbf{GLM-4.5 vs Claude Sonnet 4}: 40.4\% win, 9.6\% tie, 50.0\% loss
  \item \textbf{GLM-4.5 vs Kimi K2}: 53.9\% win, 17.3\% tie, 28.8\% loss
  \item \textbf{GLM-4.5 vs Qwen3-Coder}: 80.8\% win, 7.7\% tie, 11.5\% loss
\end{itemize}

As shown in figure~\ref{fig:successrate}, GLM-4.5 particularly excelled in tool calling reliability, achieving the highest success rate at \textbf{90.6\%}, compared to Claude Sonnet 4 (89.5\%), Kimi-K2 (86.2\%), and Qwen3-Coder (77.1\%). While Claude Sonnet 4 remains a strong competitor, GLM-4.5 outperformed other models in both task completion consistency and agentic execution robustness.

\subsubsection{Evaluation of Logical Reasoning}
To rigorously assess the true logical reasoning capabilities of the models and mitigate the risk of data contamination from common logical questions found online, we constructed a new, challenging evaluation set. This set comprises novel and complex logical reasoning problems that are structurally different from those widely available on the internet. Each problem is designed to require multiple steps of logical deduction to arrive at the correct solution.

For this evaluation, we established a unified and detailed scoring standard for each question. We then tasked each model with solving these problems. The correctness and quality of each model's response were subsequently inspected and scored by human experts. The results show a competitive landscape, with GLM-4.5 performing on par with leading models.

\begin{table}[!h]
\centering
\caption{Expert Evaluation Scores on Novel Logical Reasoning Problems}
\label{tab:logic-reasoning-results}
\begin{tabular}{lc}
\toprule
\textbf{Model} & \textbf{Score} \\
\midrule
Gemini 2.5 Pro & 65.8 \\
DeepSeek-R1-0528 & 62.1 \\
GLM-4.5 & 62.0 \\
GLM-4.5-Air & 53.4 \\
Kimi K2 & 51.9 \\
\bottomrule
\end{tabular}
\end{table}

\subsection{Evaluation of Translation}

\paragraph{The New Paradigm of Translation} Translation today extends beyond simple text conversion to encompass a nuanced understanding of evolving internet slang, cultural context, and domain-specific terminology:

\textit{Netizen Lingo:} Translating ``yyds'' accurately requires recognizing it as the acronym for the Chinese phrase \begin{CJK*}{UTF8}{gbsn}``永远的神''\end{CJK*} (yǒng yuǎn de shén), meaning ``the eternal god,'' thus capturing its true sentiment of enthusiastic praise and admiration.

\textit{Domain Nicknames:} Recognizing \begin{CJK*}{UTF8}{gbsn}``胖''\end{CJK*} (literally ``fat white'') is critical within photography communities. Specialized models may translate it incorrectly, but a general-purpose model understands it as a widely used nickname for the ``Canon EF 70-300mm f/4-5.6 IS USM'' lens, providing precise translations.

\textit{Symbols}: When a Chinese user sends a ``fish'' emoji in a conversation to refer to a second-hand marketplace, can the model understand the cultural meme behind it, which points to the \begin{CJK*}{UTF8}{gbsn}``闲鱼'' (Xiányú)\end{CJK*} platform? This tests the model's cognitive ability to connect visual symbols with online cultural phenomena.

\textit{Deep Contextual Reasoning:} Translating \begin{CJK*}{UTF8}{gbsn}``三花公主驾到，速来围观''\end{CJK*} demands identifying \begin{CJK*}{UTF8}{gbsn}``三花''\end{CJK*} not as a person's name but as a reference to the popular calico coloration of cats. A general-purpose model accurately deduces this context, translating the phrase idiomatically as ``The Calico Princess has arrived! Come and see!''.

These examples underscore modern translation as a task rooted deeply in knowledge and reasoning.

\paragraph{Evaluation Results} We tested 100 challenging, real-world cases commonly mistranslated by current tools, comparing GLM-4.5 against specialized translation models (Qwen-MT-plus, Qwen-MT-turbo, Seed-X~\cite{cheng2025seed}) in a blind human evaluation (scored 0-3 considering whether the meaning is conveyed correctly and whether the language is authentic). The results are shown in Table~\ref{tab:translation}.

\begin{table}[ht]
    \centering
    
    \caption{ Human Scores on Selected Challenging Translation data}
    \begin{tabular}{l c}
    \toprule
    Model & Average Score \\
    \midrule
    \textbf{GLM-4.5}     & \textbf{1.71} \\
    Qwen-MT-plus         & 0.38 \\
    Qwen-MT-turbo        & 0.55 \\
    Seed-X               & 0.65 \\
    \bottomrule
    \end{tabular}
    \label{tab:translation}
\end{table}

GLM-4.5 significantly outperforms specialized models. For example, translating \begin{CJK*}{UTF8}{gbsn}``三花公主驾到''\end{CJK*} specialized models failed contextually, whereas GLM-4.5 accurately conveys the idiomatic meaning.

\section{Conclusion}
In this report, we have introduced the GLM-4.5 model series, including GLM-4.5 and GLM-4.5-Air. Both models adopt the MoE architecture, which improves the computational efficiency compared to previous GLM models. GLM-4.5 excels at reasoning, coding, and agentic tasks, ranked in 3rd place globally among open-source and proprietary models. We release the model weights of GLM-4.5 and GLM-4.5-Air to advance the applications and research of large language models.
\newpage
\section{Contribution}
\label{sec:contribution}
\newcommand{\cpara}[1]{~\\\textbf{#1}~\\}
Contributors' names are listed in alphabetical order by first name. Names marked with an asterisk (*) indicate individuals who have since left our team. 

\cpara{Core Contributors}
Bin Chen, Chengxing Xie, Cunxiang Wang, Da Yin, Hao Zeng, Jiajie Zhang, Kedong Wang, Lucen Zhong, Mingdao Liu, Rui Lu, Shulin Cao, Xiaohan Zhang, Xuancheng Huang, Yao Wei, Yean Cheng, Yifan An, Yilin Niu, Yuanhao Wen, Yushi Bai, Zhengxiao Du, Zihan Wang (\begin{CJK*}{UTF8}{gbsn}汪子涵\end{CJK*}), Zilin Zhu

\cpara{Contributors}
Bohan Zhang, Bosi Wen, Bowen Wu, Bowen Xu*, Can Huang, Casey Zhao, Changpeng Cai, Chao Yu, Chen Li, Chendi Ge, Chenghua Huang, Chenhui Zhang, Chenxi Xu, Chenzheng Zhu, Chuang Li*, Congfeng Yin, Daoyan Lin, Dayong Yang, Dazhi Jiang, Ding Ai, Erle Zhu, Fei Wang, Gengzheng Pan, Guo Wang, Hailong Sun, Haitao Li, Haiyang Li, Haiyi Hu, Hanyu Zhang, Hao Peng, Hao Tai, Haoke Zhang, Haoran Wang, Haoyu Yang*, He Liu, He Zhao, Hongwei Liu, Hongxi Yan, Huan Liu, Huilong Chen, Ji Li, Jiajing Zhao, Jiamin Ren, Jian Jiao, Jiani Zhao, Jianyang Yan, Jiaqi Wang*, Jiayi Gui, Jiayue Zhao, Jie Liu, Jijie Li, Jing Li, Jing Lu, Jingsen Wang, Jingwei Yuan, Jingxuan Li, Jingzhao Du, Jinhua Du, Jinxin Liu, Junkai Zhi, Junli Gao, Ke Wang, Lekang Yang*, Liang Xu, Lin Fan, Lindong Wu, Lintao Ding, Lu Wang, Man Zhang, Minghao Li, Minghuan Xu, Mingming Zhao, Mingshu Zhai*, Pengfan Du, Qian Dong, Shangde Lei, Shangqing Tu, Shangtong Yang, Shaoyou Lu, Shijie Li, Shuang Li (\begin{CJK*}{UTF8}{gbsn}李泷\end{CJK*}), Shuang Li (\begin{CJK*}{UTF8}{gbsn}李爽\end{CJK*}), Shuxun Yang, Sibo Yi*, Tianshu Yu, Wei Tian, Weihan Wang, Wenbo Yu, Weng Lam Tam, Wenjie Liang, Wentao Liu, Xiao Wang*, Xiaohan Jia, Xiaotao Gu, Xiaoying Ling, Xin Wang, Xing Fan, Xingru Pan, Xinyuan Zhang, Xinze Zhang, Xiuqing Fu, Xunkai Zhang, Yabo Xu, Yandong Wu, Yida Lu, Yidong Wang, Yilin Zhou, Yiming Pan, Ying Zhang, Yingli Wang, Yingru Li, Yinpei Su, Yipeng Geng, Yitong Zhu, Yongkun Yang*, Yuhang Li, Yuhao Wu*, Yujiang Li, Yunan Liu, Yunqing Wang, Yuntao Li, Yuxuan Zhang, Zezhen Liu, Zhen Yang, Zhengda Zhou, Zhongpei Qiao, Zhuoer Feng, Zhuorui Liu, Zichen Zhang, Zihan Wang (\begin{CJK*}{UTF8}{gbsn}王梓汉\end{CJK*}), Zijun Yao, Zikang Wang, Ziqiang Liu, Ziwei Chai, Zixuan Li, Zuodong Zhao*

\cpara{Tech Leads}
Aohan Zeng, Xin Lv, Qinkai Zheng, Zhenyu Hou

\cpara{Advisors}
Jie Tang, Yuxiao Dong, Juanzi Li, Hongning Wang, Minlie Huang, Bin Xu, Jidong Zhai, Wenguang Chen

\cpara{Acknowledgement}
We are grateful for all the support from Beijing, Shanghai, Tianjin, Hangzhou, Zhuhai, and Chengdu.
Special thanks to our customers and community developers.

\clearpage

\bibliographystyle{abbrv}
\bibliography{ref}

\clearpage

\end{document}